\documentclass[letterpaper]{article} 
\usepackage{aaai2026}  
\usepackage{times}  
\usepackage{helvet}  
\usepackage{courier}  
\usepackage[hyphens]{url}  
\usepackage{graphicx} 
\usepackage{natbib}  
\usepackage{caption} 
\usepackage{algorithm}
\usepackage{algpseudocode}
\usepackage{amsfonts}

\usepackage{amsmath}
\DeclareMathOperator*{\argmax}{arg\,max}

\usepackage{booktabs}    
\usepackage{multirow}    
\usepackage{newfloat}
\usepackage{listings}
\DeclareCaptionStyle{ruled}{labelfont=normalfont,labelsep=colon,strut=off} 
\floatstyle{ruled}
\newfloat{listing}{tb}{lst}{}
\floatname{listing}{Listing}
\usepackage{amsmath}
\usepackage{amsthm}
\usepackage{color}
\usepackage[table]{xcolor}
\usepackage{makecell}
\usepackage{enumitem}
\usepackage{subfigure}
\usepackage{subcaption}
\usepackage{longtable}
\usepackage{pifont}
\usepackage{placeins}

\theoremstyle{plain} 

\theoremstyle{definition} 

\theoremstyle{remark} 

\usepackage{mdframed}
\usepackage{xcolor}
\usepackage{lipsum}

\title{OptScale: Probabilistic Optimality for Inference-time Scaling}

\author{%
Youkang Wang$^{1,2 *}$, ~~Jian Wang$^2$\thanks{Equal contribution.}, ~~Rubing Chen$^{1,2}$, ~~Xiao-Yong Wei$^{3,1,2}$\footnote{Corresponding author.}
}

\affiliations{
$^1$ PolySmart Group  ~~ $^2$ Department of Computing, The Hong Kong Polytechnic University \\
$^3$ College of Computer Science, Sichuan University \\
\{yk2022.wang, rubing.chen\}@connect.polyu.hk \\
\{cs007.wei\}@polyu.edu.hk
}

\begin{document}

\maketitle

\begin{abstract}
Inference-time scaling has emerged as a powerful technique for enhancing the reasoning performance of Large Language Models (LLMs). However, existing approaches often rely on heuristic strategies for parallel sampling, lacking a principled foundation. 
To address this gap, we propose a probabilistic framework that formalizes the optimality of inference-time scaling under the assumption that parallel samples are independently and identically distributed (i.i.d.), and where the Best-of-$N$ selection strategy follows a probability distribution that can be estimated. Within this framework, we derive a theoretical lower bound on the required number of samples to achieve a target performance level, providing the first principled guidance for compute-efficient scaling.
Leveraging this insight, we develop \textsc{OptScale}, a practical algorithm that dynamically determines the optimal number of sampled responses. \textsc{OptScale} employs a language model-based predictor to estimate probabilistic prior parameters, enabling the decision of the minimal number of samples needed that satisfy predefined performance thresholds and confidence levels.
Extensive experiments on representative reasoning benchmarks (including MATH-500, GSM8K, AIME, and AMC) demonstrate that \textsc{OptScale} significantly reduces sampling overhead while remaining better or on par with state-of-the-art reasoning performance. Our work offers both a theoretical foundation and a practical solution for principled inference-time scaling, addressing a critical gap in the efficient deployment of LLMs for complex reasoning.
%

\end{abstract}

\begin{links}
\link{Code}{https://github.com/Albertwyk/OptScale}
\end{links}

\section{Introduction}

The reasoning capabilities of LLMs have become a pivotal research area, given their ability to tackle complex cognitive tasks such as mathematical reasoning, decision-making, and problem-solving \cite{zhang2025and, yang2025towards}. 
Recent advances demonstrate that inference-time scaling can substantially enhance reasoning performance by enabling the model to generate diverse candidate solutions, creating a richer space for comprehensive inference through answer aggregation \cite{chen2025seal,ding2025dynamic, wan2024dynamic}. 
This approach is particularly valuable for proprietary LLMs, as it can operate purely at inference time without requiring model modifications.
Although effective, these techniques face a fundamental efficiency challenge: the linear relationship between candidate solutions and computational costs leads to prohibitive token consumption. 
This has created an urgent need for optimized scaling strategies that navigate the Pareto frontier between reasoning performance and computational efficiency, maximizing accuracy while minimizing redundant computation \cite{qu2025optimizing}.

Parallel inference-time scaling, a dominant paradigm in this space, operates by generating $N$ candidate responses in parallel and selecting the optimal output via learned rules (e.g., verifiers) \cite{prmlessons} or consensus (e.g., majority voting) \cite{wang2023self}.
Current approaches fall into three categories: 
verifier-based approaches that train auxiliary models to rerank candidates \cite{yang2025towards}; data-driven methods that fine-tune LLMs on high-quality reasoning traces to improve solution quality \cite{qu2025optimizing}; and inference-time techniques that dynamically adjust the number of candidates per input \cite{chen2025seal}.
While these methods demonstrate practical effectiveness, they suffer from significant theoretical limitations. 
They rely heavily on empirical heuristics and implicit priors (such as learned verifier preferences or empirical token budgets) without proper mathematical formulation. 
Crucially, the field lacks a systematic derivation of these priors from first principles or a rigorous quantification of the fundamental efficiency-accuracy trade-offs inherent in parallel scaling.

This paper presents a foundational study that bridges this critical gap through three key contributions:
First, we formulate a probabilistic framework that formally establishes the optimality conditions for parallel inference-time scaling under i.i.d. assumptions, where the Best-of-N selection process follows an estimable probability distribution. 
This theoretical formulation provides a principled understanding of scaling behavior in probabilistic terms.
Second, we derive a theoretical lower bound on the number of samples required to achieve any target performance level. 
This result offers mathematically grounded guidance for compute-efficient scaling, representing a significant advance beyond current heuristic approaches.
Third, we develop \textsc{OptScale}, an efficient algorithm that implements these theoretical insights in practice. \textsc{OptScale} dynamically determines the optimal sample size by employing an LLM-based predictor to estimate prior parameters, ensuring minimal computational cost while satisfying predefined performance thresholds and confidence levels.
Our experiments demonstrate \textsc{OptScale}'s computational efficiency: it automatically terminates sampling for simple questions while preventing excessive computation on intractable ones.

To our knowledge, this work presents the first comprehensive treatment of parallel inference-time scaling that: 1) establishes a rigorous probabilistic model of the scaling process, 2) derives fundamental limits on scaling efficiency, and 3) provides a practical algorithmic implementation that approaches these theoretical optima.
\section{Related Work}

\subsection{LLM Reasoning}
The ability of LLMs to perform complex reasoning tasks has garnered significant attention and seen rapid advancements \cite{yu2024natural, xu2025towards}. Early successes with prompting strategies like Chain-of-Thought (CoT) \cite{wei2022chain} demonstrated that LLMs could produce step-by-step reasoning traces, significantly improving performance on tasks requiring logical deduction, mathematical problem-solving, and multi-step planning \cite{ahn2024large, snell2024scaling, liang2024improving}. This foundational work has spurred the development of more sophisticated prompting techniques, including Self-Consistency \cite{wan2024dynamic}, which generates multiple reasoning paths and selects the most consistent answer, and Tree-of-Thought (ToT) reasoning, which allows models to explore diverse reasoning trajectories in parallel \cite{yao2023tree, long2023large}.

\subsection{Parallel Inference-time Scaling}
Inference-time scaling \cite{ke2025survey} has emerged as a key strategy to enhance LLM reasoning by enlarging computational resources during inference. In this paper, we primarily focus on parallel inference-time scaling, which involves generating multiple candidate solutions in parallel and selecting the best one \cite{snell2024scaling}.

\paragraph{Verifier-based Scaling Methods.}
To improve inference-time performance, many studies use verifiers, also known as process reward models (PRMs), to assess the quality of candidate solutions generated by a primary LLM \cite{setlur2024rewarding, qu2025optimizing}. These verifiers predict the correctness of a solution or assign it a quality score, guiding the selection of the final answer by re-ranking the $N$ outputs. For example, some methods train verifiers using human feedback or generated labels \cite{bai2022training}. Recent work examines how the interaction between generators, PRMs, and problem difficulty affects optimal test-time scaling \cite{prmlessons}. However, relying on additional models increases computational overhead, and the strategies for determining $N$ often lack a theoretical foundation.

\paragraph{Dynamic Scaling Methods.}
Recognizing the limitations of fixed $N$, several methods explore dynamic approaches to adjust computation at inference time. These techniques adapt the number of generated candidates or computational resources based on the input query or the ongoing generation process \cite{ding2025dynamic, wan2024dynamic}. For example, \citet{chen2025seal} allocates more computation to complex problems or where initial candidate solutions show high uncertainty or disagreement. Approaches like dynamic decomposition break down solution paths into manageable steps, dedicating more resources to challenging sub-problems. The goal is to avoid unnecessary computation on simpler instances while providing sufficient resources for harder ones. While these dynamic methods improve efficiency, they often rely on heuristic rules or learned strategies without a clear theoretical framework for the trade-off between efficiency and accuracy. These dynamic methods represent a step towards more efficient scaling. 

However, the rules governing the dynamic allocation are often heuristic or learned without an explicit efficiency-accuracy trade-off. Our work distinguishes itself by providing a principled probabilistic formulation of the parallel scaling. This allows us to derive a provably optimal strategy for dynamically allocating compute to maximize accuracy per token and to quantify the theoretical upper bound for efficiency, addressing the lack of systematic, theoretically grounded priors in current dynamic scaling approaches. 
\section{Scaling with Probabilistic Optimality}
\label{sec:method}

In this section, we formulate a theoretical framework for inference-time scaling from a probabilistic optimality perspective. Based on this framework, we present \textsc{OptScale}, an implementation that effectively achieves the goal of compute-efficient inference-time scaling.

\subsection{Theoretical Framework}
\label{ssec:theory_framework}

\paragraph{Preliminaries.}

Given an input question $q$ processed by an LLM $\mathcal{M}$, which generates $N$ candidate answers $\{a_i\}_{i=1}^N$.
Each answer $a_i$ consists of $T_i$ reasoning steps $\{a_{i,t}\}_{t=1}^{T_i}$. 
A verifier $\mathcal{V}$ (e.g., a Process Reward Model, PRM) assigns normalized scores $\{s_{i,t}\in [0,1]\}_{t=1}^{T_i}$ to each step.
Defining $\mathcal{A}(\cdot)$ as an aggregation operator (typically averaging multi-step scores or using the final step score), we compute the overall verification score for answer $a_i$ as:
\begin{equation}
s_i = \mathcal{A} \big[\,\mathcal{V}(a_{i,t}\,\vert\,q,\mathcal{M},a_{i,{<t}})\,\big]
\end{equation}
The optimal answer $a^{*}$ is selected through:
\begin{equation}
a^* = \argmax_{i \in \{1,\ldots,N\}} s_i
\label{eq:max_answer}
\end{equation}
Previous research on inference-time reasoning has primarily differed in their choices of the language model $\mathcal{M}$, verifier $\mathcal{R}$, and aggregator $\mathcal{A}$.
In contrast, our work makes a fundamental shift in perspective by investigating the probabilistic distributions that govern these components. 
It is a crucial distinction that sets our approach apart from prior studies.

\paragraph{Verifier Score Distribution.}
Let us assume that the verifier scores $\{s_i\}$ for a fixed question $q$ are samples for a random variable $S$ that is conform to a continuous probability distribution with probability density function (PDF) $f_S(\cdot)$ and cumulative distribution function (CDF) $F_S(\cdot)$ as:
\begin{align}
S\sim f_S(s|\theta,q),\, F_S(s)=\mathbb{P}(S<s)=\int_{-\infty}^sf_S(x)dx,
\end{align}
where $\theta$ represents the parameters in $\mathcal{M}$, $\mathcal{R}$, and hyperparameters for generation (e.g., the temperature $t$).

\paragraph{Distribution of the Maximum Verifier Score.} 

Let $Y = \max\{s_1, s_2,..., s_N\}$ denote the maximum verification score among $N$ candidates, which serves as the core selection criterion in Eq.~(\ref{eq:max_answer}). 
For any observed score $s$, the optimal decision on whether to sample more candidates can be determined when we know the exceedance probability
$\mathbb{P}(Y \leq s)$.
This represents the likelihood that $s$ is the maximal score in the complete population of possible verifications.
The probability can be derived by considering the joint event where $s_i\leq s$ holds for all $N$ independent draws, yielding:
\begin{equation}
\mathbb{P}(Y \leq s) = \prod_{i=1}^N \mathbb{P}(s_i \leq s)= [F_S(s)]^N,
\label{eq:CDF_Y}
\end{equation}
which also defines the CDF of $Y$ as $F_Y(s)=\mathbb{P}(Y \leq s)$.
The PDF of $Y$ is:
\begin{equation}
f_Y(s) = \frac{d}{ds}F_Y(s) = N[F_S(s)]^{N-1}f_S(s).
\end{equation}
Note that both the PDF and the CDF of $Y$ are functions with respect to $N$. Figure \ref{fig:verifier_score_distribution} shows a few examples of such PDF functions in different values of $N$.

\paragraph{Probabilistic Optimality.}
After $N$ sampling rounds, conventional methods (e.g., Best-of-$N$) confront a critical trade-off: whether to continue sampling (potentially finding higher-scoring answers) or terminate (to conserve computational resources). 
Our solution derives the probabilistically optimal sample size 
$N^*$ as the minimal value satisfying:
\begin{equation}
\mathbb{P}(Y \geq s_{\min}) \geq \alpha
\label{eq:stopping_criteria}
\end{equation}
where $s_{\min}$ denotes the quality threshold for valid solutions, $\alpha$ represents the required confidence level.
From Eq.~(\ref{eq:CDF_Y}), we reformulate the probability requirement as:
\begin{equation}
1 - F_Y(s_{\min}) \geq \alpha \implies [F_S(s_{\min})]^N \leq 1-\alpha.
\end{equation}
Solving for $N$ yields the following closed-form solution:
\begin{equation}
N^* \geq \left\lceil \frac{\log(1-\alpha)}{\log F_S(s_{\min})} \right\rceil,
\label{eq:opt_N_theo}
\end{equation}
where $\lceil \cdot \rceil$ ensures integer sample sizes.

\subsection{Implementation of the Optimal Scaling}
\label{ssec:detailed_method}

Translating the theoretical framework into practice raises a critical question: \textit{how can we reliably estimate the verifier score distribution $f_S(s|\theta,q)$?}
Let $(\mu, \sigma)$ parameterize this distribution, where these parameters need not imply Gaussianity.
While direct estimation of $(\mu, \sigma)$ is non-trivial, we find that $f_S(s|\theta,q)$ can be modeled as a \textbf{truncated normal distribution} constrained to $[0,1]$. The probability density function is given by:
\begin{equation}
f_S(s|\theta,q) \propto f_S(s|\mu,\sigma) = \frac{\phi(\frac{s-\mu}{\sigma})}{\sigma[\Phi(\frac{1-\mu}{\sigma}) - \Phi(\frac{0-\mu}{\sigma})]},
\label{eq:f_hypo}
\end{equation}
where $\phi(\cdot)$ and $\Phi(\cdot)$ denote the standard normal PDF and CDF. 
The optimal sample size $N^*$ is derived via Eq.~(\ref{eq:opt_N_theo}) using $(\mu, \sigma)$ estimates.
To this end, we propose \textsc{OptScale} with the following two alternatives to implement inference-time scaling with probabilistic optimality.

\paragraph{\textsc{OptScale}$^t$: Parameter Estimation via Trainable Predictors.} 
This approach uses offline data to learn global trends in $(\mu, \sigma)$. 
We model them as standard Gaussian random variables centered around predicted means $(\bar\mu, \bar\sigma)$:
\begin{equation}
\mu\sim \mathcal{N}(\bar\mu,\sigma_\mu), \quad \sigma\sim \mathcal{N}(\bar\sigma,\sigma_\sigma),
\label{eq:mean_distribution}
\end{equation}
where $\sigma_\mu$, $\sigma_\sigma$ control estimation error bounds. 
The trends $(\bar\mu, \bar\sigma)$ are obtained by predictors via fine-tuning MLPs:
\begin{equation}
\bar{\mu} = \text{MLP}_{\mu}(q, \mathcal{M}),  \quad \bar{\sigma} = \text{MLP}_{\sigma}(q, \mathcal{M}).
\end{equation}
Given observed verifier scores $D=\{s_k\}$ at inference, we refine $(\bar\mu, \bar\sigma)$ via Maximum-a-Posteriori (MAP) estimation:
\begin{align}
\mu^*, \sigma^* = \argmax_{\mu,\sigma} \Big(& \!\sum_{k=1}^{|D|} \log f_{S}(s_k|{\mu},{\sigma}) + \log f_{\mu}(\mu|\bar\mu,\sigma_\mu)  \notag\\  & + \log f_{\sigma}(\sigma|\bar\sigma,\sigma_\sigma)\!\Big).
\end{align}
where $f_\mu$, $f_\sigma$ represent prior densities of $\mu$ and $\sigma$ in Eq.~(\ref{eq:mean_distribution}).

\paragraph{\textsc{OptScale}$^0$: Training-free Parameter Estimation.}
This variant estimates parameters solely from observed data using Maximum Likelihood Estimation (MLE), reducing the need for additional training.
For a new query $q$, we empirically initialize $(\mu_0, \sigma_0)$ using a heuristic strategy with uniform uncertainty. We then use bootstrapping to calculate $\mu$ and $\sigma$ from previous scores iteratively.
As more verifier scores $\{s_k\}$ are observed, we update the parameters by:
\begin{equation}
\mu^*, \sigma^* = \argmax_{\mu,\sigma} \sum_{k=1}^{|D|} \log f_S(s_k|{\mu},{\sigma}).
\end{equation}
This variant does not rely on learned predictors, making it highly lightweight and efficient.

\paragraph{Adaptive Scaling with \textsc{OptScale}.} 

Both variants ultimately compute the optimal sample size $N^{*}$ using their refined parameters $(\mu^{*}, \sigma^{*})$ in Eq. (\ref{eq:opt_N_theo}), which depends on the tail distribution $F_S(s_{\min})$.
Sampling continues until the current sample count $N \geq N^{*}$, enabling 1) early stopping for simple queries, and 2) bounded effort for more complex or ambiguous cases.
This adaptive mechanism allows \textsc{OptScale} to efficiently scale inference-time computation.
\section{Experiments}

\begin{figure*}[t!]
    \centering
    \includegraphics[width=1\linewidth]{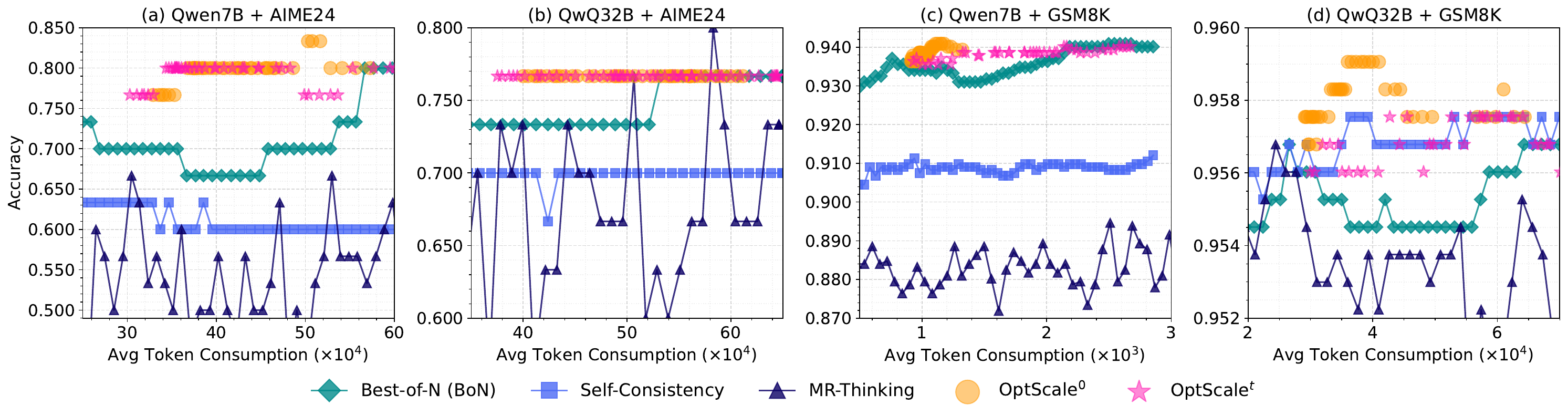}
    \caption{Scaling efficiency comparisons (accuracy vs. average token consumption): Both \textsc{OptScale}$^0$ and \textsc{OptScale}$^t$ show consistently faster convergence and optimal accuracy-token tradeoff over the compared baseline methods.}
    \label{fig:scale_efficiency}
\end{figure*}

\begin{figure*}[t!]
    \centering
    \includegraphics[width=1\linewidth]{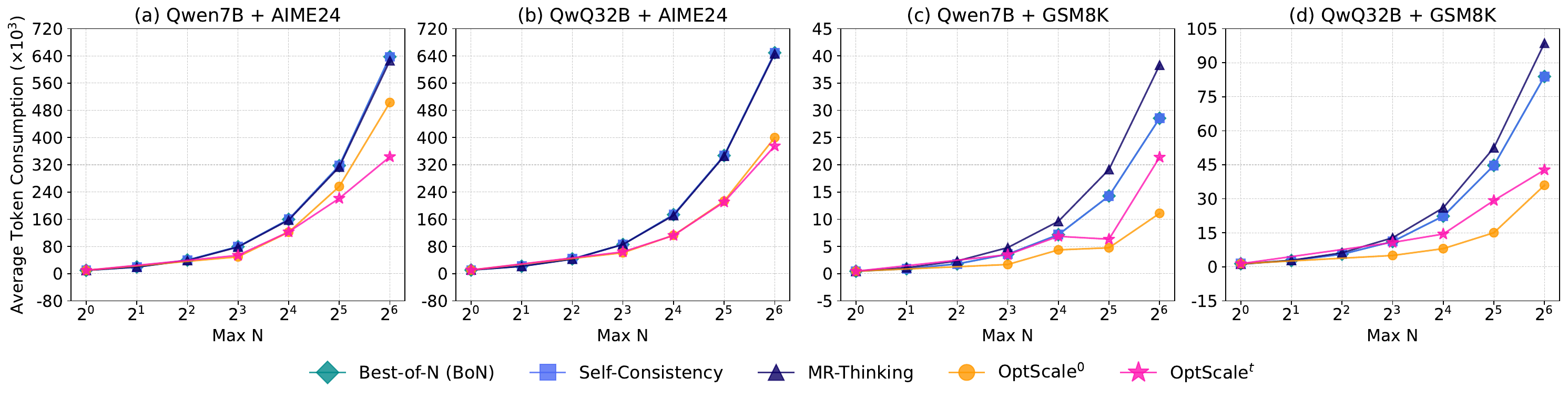}
    \caption{Token consumption of different methods when scaling across $N$: \textsc{OptScale} consistently achieves reduced completion tokens when scaling to higher $N$ over the compared baseline methods.}
    \label{fig:scale_n_results}
\end{figure*}

\begin{table*}[t!]
\centering
\resizebox{0.98\textwidth}{!}{%
\begin{tabular}{l c c c c c c c c c c}
\toprule
\multirow{2}{*}{\textbf{Baseline Method}} & \multicolumn{2}{c}{\textbf{MATH-500}} & \multicolumn{2}{c}{\textbf{GSM8K}} & \multicolumn{2}{c}{\textbf{AIME 2024}} & \multicolumn{2}{c}{\textbf{AIME 2025}} & \multicolumn{2}{c}{\textbf{AMC 2023}} \\ 
\cmidrule{2-11}
 & \textbf{Acc.}  & \textbf{Toks.} ($\downarrow$) & \textbf{Acc.} & \textbf{Toks.} ($\downarrow$) & \textbf{Acc.}  & \textbf{Toks.} ($\downarrow$)  & \textbf{Acc.} & \textbf{Toks.} ($\downarrow$) & \textbf{Acc.}  & \textbf{Toks.} ($\downarrow$) \\
\midrule
\multicolumn{11}{l}{\textbf{Deepseek-R1-Distill-Qwen-7B}} \\ 
\midrule
Best-of-N (BoN) ($N=8$)& \textbf{94.8}& 22135& \textbf{92.4}& 3582& \textbf{70.0}& 79367& 43.3& 84342& \textbf{95.0}& 40511
\\
Self-Consistency ($N=8$)& 93.4& 22135& 90.1& 3582& 60.0& 79367& 40.0& 84342& 85.0& 40511
\\
 MR-Thinking ($N=8$)& 91.2& 21396& 88.4& 4792& 56.7& 78432& 40.0& 86568& 87.5&36780
\\
\textsc{OptScale}$^0$ (Ours) ($N=8$)& \textbf{94.8}& \textbf{11354}& \textbf{92.4}& \textbf{1687}& \textbf{70.0}& \textbf{49505}& 43.3& 78803& \textbf{95.0}& \textbf{29288}\\
\textsc{OptScale}$^t$  (Ours) ($N=8$)& \textbf{94.8}& 18236& \textbf{92.4}& 3492& \textbf{70.0}& 53855& \textbf{46.7}& \textbf{69661}& \textbf{95.0}&30671
\\
\midrule
Best-of-N (BoN) ($N=64$)& 94.0& 174693& 94.0& 28547&80.0& 637293& \textbf{53.3}&  676533& \textbf{95.0}& 312241
\\
 Self-Consistency ($N=64$)& 93.4& 174693& 91.2& 28547& 60.0& 637293& 40.0& 676533& 92.5&312241
\\
 MR-Thinking ($N=64$)& 92.0& 168331& 88.2& 38305& 70.0& 625850& 40.0& 684446& 90.0&286639
\\
 \textsc{OptScale}$^0$ (Ours) ($N=64$)& \textbf{94.6}& 110001& \textbf{94.1}& \textbf{11086}& \textbf{83.3}& 503002& 50.0& \textbf{549344}& \textbf{95.0}&119777
\\
 \textsc{OptScale}$^t$ (Ours) ($N=64$)& \textbf{94.6}& \textbf{76284}& 94.0& 21386& 80.0& \textbf{343491}& \textbf{53.3}& 649900& \textbf{95.0}&\textbf{119282}\\
\midrule
\midrule
\multicolumn{11}{l}{\textbf{Llama-3.1-8B-Instruct}} \\
\midrule
Best-of-N (BoN) ($N=8$)
& 63.6& 9609& \textbf{88.4}& 2160& \textbf{10.0}& 54212& \textbf{3.3}& 51510& \textbf{32.5}& 15321
\\
Self-Consistency ($N=8$)
& 58.6& 9609& 87.5& 2160& 3.3& 54212& 0.0& 51510& 27.5& 15321
\\
MR-Thinking ($N=8$)& 41.6& 14918& 61.6& 2812& 0.0& 46037& \textbf{3.3}& \textbf{39374}& 12.5& 24814
\\
\textsc{OptScale}$^0$  (Ours) ($N=8$)
& \textbf{63.8}& \textbf{8756}& \textbf{88.4}& \textbf{1462}& \textbf{10.0}& \textbf{42859}& \textbf{3.3}&  46970& \textbf{32.5}& 15113
\\
\textsc{OptScale}$^t$ (Ours) ($N=8$)& 63.6& 9479& \textbf{88.4}& 2136& \textbf{10.0}& 43212& \textbf{3.3}& 46724& \textbf{32.5}&\textbf{14323}\\
\midrule
 Best-of-N (BoN) ($N=64$)
& 68.8& 70643& 89.3& 14697& \textbf{13.3}& 386783& \textbf{0.0}& 348361& 42.5&113401
\\
Self-Consistency ($N=64$)
& 60.8& 70643& 89.2& 14697& 6.7& 386783& \textbf{0.0}& 348361& 32.5& 113401
\\
 MR-Thinking ($N=64$)& 40.8& 122227& 60.7& 23423& 0.0& \textbf{377475}& \textbf{0.0}& 327074& 15.0&189239
\\
 \textsc{OptScale}$^0$  (Ours) ($N=64$)
& \textbf{69.0}& 64179& 89.5& \textbf{5720}& \textbf{13.3}& 386772& \textbf{0.0}& 141904& \textbf{45.0}&\textbf{100675}\\
 \textsc{OptScale}$^t$ (Ours) ($N=64$)& 68.6& \textbf{60692}& \textbf{89.6}& 12044& \textbf{13.3}& 386668& \textbf{0.0}& \textbf{140707}& \textbf{45.0}&106039
\\
 \midrule
 \midrule
\multicolumn{11}{l}{\textbf{Qwen/QwQ-32B}} \\
\midrule
Best-of-N (BoN) ($N=8$)
& 94.6& 30601& 95.8& 11177& \textbf{73.3}& 85853& \textbf{66.7}& 95235& 95.0& 55291
\\
Self-Consistency ($N=8$)
& \textbf{95.2}& 30601& 95.6& 11177& 70.0 & 85853& \textbf{66.7}& 95235& 92.5& 55291
\\
MR-Thinking ($N=8$)& 94.6& 30378& 95.3& 12832& 70.0& 85864& 53.3& 95536& 87.5& 47288
\\
\textsc{OptScale}$^0$  (Ours) ($N=8$)
& 95.0& \textbf{17469}& \textbf{95.9}& \textbf{5031}& \textbf{73.3}& \textbf{61449}& \textbf{66.7}&  85669& \textbf{95.0}& 42177
\\
\textsc{OptScale}$^t$  (Ours) ($N=8$)& 94.8& 26910& 95.8& 10767& \textbf{73.3}& 64094& \textbf{66.7}& \textbf{84766}& \textbf{95.0}&\textbf{36494}
\\
\midrule
 Best-of-N (BoN) ($N=60$)
& 94.8& 230402& 95.8& 83902& \textbf{76.7}& 649256& 66.7& 721405& 97.5&420481
\\
Self-Consistency ($N=60$)
& 95.4& 230402& 95.8& 83902& 70.0& 649256& 63.3& 721405& 92.5& 420481
\\
 MR-Thinking ($N=60$)& 93.8& 227838& 94.8& 98607& 73.3& 646201& 56.7& 719654& 80.0&360068
\\
 \textsc{OptScale}$^0$  (Ours) ($N=60$)
& \textbf{95.8}& 107720& \textbf{95.9}& \textbf{35985}& \textbf{76.7}& 399929& \textbf{70.0}& 556340& \textbf{100.0}&\textbf{190633}
\\
 \textsc{OptScale}$^t$ (Ours) ($N=60$)& \textbf{95.8}& \textbf{106412}& 95.8& 42735& \textbf{76.7}& \textbf{375208}& \textbf{70.0}& \textbf{516346}& \textbf{100.0}&202603
\\

\bottomrule
\end{tabular}
}
\caption{Comparison of different inference-time scaling methods on common mathematical reasoning benchmarks (with $N=8$ and $N=64$). ``Acc.'' denotes accuracy (\%), ``Toks.'' indicate the total number of inference tokens.}
\label{tab:sota_results}
\end{table*}

\subsection{Experimental Setup}
\label{sec:experiment_setup}

\paragraph{Benchmarks.} 
We employ the following representative reasoning benchmarks: 
(1) \textbf{MATH-500}~\cite{hendrycks2measuring} with 500 high school competition problems across algebra and geometry, following the OpenAI evaluation split; (2)
\textbf{GSM8K}~\cite{cobbe2021training} with 8.5K grade school word problems assessing basic arithmetic and textual reasoning; (3)
\textbf{AIME 2024}~\cite{maa2024aime} and (4) \textbf{AIME 2025}\footnote{\url{https://huggingface.co/datasets/math-ai/aime25}.}, with each containing 30 pre-Olympiad level problems from the American Invitational Mathematics Examination, designed to test advanced mathematical reasoning; (5) \textbf{AMC}\footnote{\url{https://huggingface.co/datasets/math-ai/amc23}.}, a collection of intermediate-level problems from the 2023 American Mathematics Competitions. We report accuracy and tokens as evaluation metrics~\cite{chongwahngo2007trecvid, chongwahngo2008trecvid}.

\paragraph{Backbone Models.}

To assess generalizability and efficiency, we employ the following open-source backbone LLMs with different sizes and architectures: 
1) \textbf{DeepSeek-R1-Distill-Qwen-7B}~\cite{guo2025deepseek}: a 7B-parameter reasoning model distilled from larger DeepSeek R1 models into a Qwen model, representing a state-of-the-art distilled model optimized for complex reasoning.
2) \textbf{Llama-3.1-8B-Instruct}~\cite{grattafiori2024llama}: an instruct model based on Llama architecture;
3) \textbf{QwQ-32B}~\cite{yang2025qwen3}: a high-capacity model specialized for complex reasoning tasks.
4) \textbf{Deepseek-R1-Distill-Llama-8B}\footnote{Main results of 1), 2), and 3) are shown in subsequent sections, while results of 4), 5), and 6) together with results of various $N$ settings for all models are showcased in the Appendix. } ~\cite{guo2025deepseek}:  an 8B reasoning model distilled from larger DeepSeek R1 models into Llama.
5) \textbf{Qwen3-8B}\footnotemark[\value{footnote}]~\cite{yang2025qwen3}, Qwen's latest reasoning model; 
6) \textbf{Llama-3.2-1B-Instruct}\footnotemark[\value{footnote}]~\cite{meta2024llama3}, a light-weighted instruction-following model in Llama series.

\paragraph{Baseline Methods.}
We benchmark \textsc{OptScale} against the commonly used inference-time scaling methods with different max $N$ settings, including: 
1) \textbf{Best-of-$N$ (BoN)}: It samples $N$ reasoning paths, then employs a verifier for scoring, and ultimately selects the best answer with the highest score;
2) \textbf{Self-Consistency (SC)}~\cite{wang2023self}: It samples multiple diverse responses and aggregates the final answer based on majority voting;
and 3) \textbf{Multi-round Thinking (MR-Thinking)}~\cite{tian2025think}: It iteratively refines the reasoning process through multiple rounds.

\paragraph{Implementation Details.}
\label{sec:implement_details}

For scoring, we use Qwen2.5-Math-PRM-7B as our verifier. For both \textsc{OptScale}$^t$ and \textsc{OptScale}$^0$, initial sample sizes of generated completions for enabling MLE/MAP refinement are set to be around 1/3 of the maximum allowed number of $N$ generations. Across all experiments for \textsc{OptScale}$^0$, we initialize $\mu_0$ and $\sigma_0$ as the mean and standard deviation of the initial sampled data for each new query.
To ensure a fair comparison, we re-ran all baseline methods under the same experimental conditions. For all datasets, we use a sampling temperature of 0.7 and a top-$p$ of 0.95. All experiments are conducted on a single NVIDIA A6000 server with 8 GPUs. Further details are presented in the Appendix.
We report accuracy (\textbf{Acc.}) and completion tokens (\textbf{Toks.}) as evaluation metrics.

\paragraph{Details on Distribution Predictor in \textsc{OptScale}$^t$.} 
Since \textsc{OptScale}$^t$ employs an auxiliary predictor for estimating the mean and standard deviation of the verifier score distribution, we introduce its training details in this section.
We use Deepseek-R1-Distill-Qwen-1.5B as the backbone model to build the distribution predictor. We then freeze all layers except the last two and train the two individual MLP layers. We curate the training dataset by taking the MATH training set, which contains approximately 4,500 questions, and pre-generating 60 completions per question. 
This process allows us to obtain the verifier score statistics and create the $<$question, mean$>$ and $<$question, standard deviation$>$ pairs, accordingly. 
We train the predictor for 30 epochs, with a learning rate of 1e-5, weight decay of 0.01, and gradient clipping at 0.1. We set a dropout ratio of 0.2 for all MLP layers during training.

\subsection{Can \textsc{OptScale} Ensure Optimal Scaling?}
To evaluate scaling efficiency, we conducted comprehensive experiments across 5 benchmark datasets and 6 backbone models, varying the maximum number of answer candidates from $1$ to $64$. 
This experimental design required 1920 total runs per method ($5$ datasets $\times$ $6$ backbones $\times$ $64$ candidate limits), representing a significant computational investment. 
Due to resource constraints, we focus our comparison on the 3 most representative baseline methods: Best-of-$N$ (BoN), Self-Consistency (SC), and Multi-Round Thinking.

Figure \ref{fig:scale_efficiency} shows that \textsc{OptScale} achieves optimal scaling. We take Deepseek-R1-Distill-Qwen-7B and QwQ-32B on GSM8K and AIME24 to evaluate both \textsc{OptScale}$^0$ and \textsc{OptScale}$^t$'s scaling capacity. We examined different quality thresholds, ranging from 0.9 to 0.99, and confidence levels, from 0.9 to 0.98.
Notably, all \textsc{OptScale} implementations along the frontier consistently outperform baselines, delivering higher accuracy with lower computational cost (token consumption). This confirms its ability to maintain scaling optimality across configurations.
Figure \ref{fig:scale_n_results} shows the scaling progress of different baseline models across the maximum allowed $N$ over the values $\{1,2,4,8,16,32,64\}$. It is clearly demonstrated that \textsc{OptScale} consistently reduces tokens compared to most baseline methods. 

These results reveal several key insights:
1) \textsc{OptScale} consistently outperforms all baseline methods across different configurations, achieving either higher accuracy with comparable token consumption or similar accuracy with substantially fewer tokens. On Qwen7B + AIME24, \textsc{OptScale} constantly has 7\% - 14\% accuracy advantage over the best-performing baseline given the same token levels, while on QwQ32B + AIME24 and Qwen7B + GSM8K, \textsc{OptScale} reaches best accuracy using 27\% and 51\% fewer tokens, exhibiting strong early convergence capabilities. 
2) \textsc{OptScale}'s superiority is particularly apparent with larger $N$ value. \textsc{OptScale}'s average token reduction percentage compared to the best baseline is 38\%, 47\%, and 54\% when $N=16,32,64$ respectively, indicating \textsc{OptScale}'s increased efficiency on larger $N$ values. We also observe that token reduction is more apparent on easier benchmarks, where QwQ32B + GSM8K achieves the highest average token reduction percentage of 64\%, while accuracy gains are greater on harder benchmarks (e.g., AIME24), proving our adaptive strategy’s strength in complex, variable-difficulty tasks.
3) Both \textsc{OptScale}$^0$ and \textsc{OptScale}$^t$ consistently consume significantly fewer tokens than other methods at the same accuracy levels, while \textsc{OptScale}$^0$ is usually more efficient on easier benchmarks (like GSM8K) and \textsc{OptScale}$^t$ saves more tokens for harder benchmarks (like AIME24).
4) Through proper dynamic allocation of inference-time compute, \textsc{OptScale} sometimes achieves accuracy levels that BoN fails to reach even when scaled indefinitely to a substantial amount of tokens, such as reaching the accuracy of 95.9\% on QwQ32B + GSM8K. These advantages stem primarily from \textsc{OptScale}'s ability to evaluate question difficulty in advance and allocate resources optimally through our probabilistic framework. Further elaboration can be found in the case studies within the Appendix.

\begin{figure}[t!]
    \centering
    \includegraphics[width=1\linewidth]{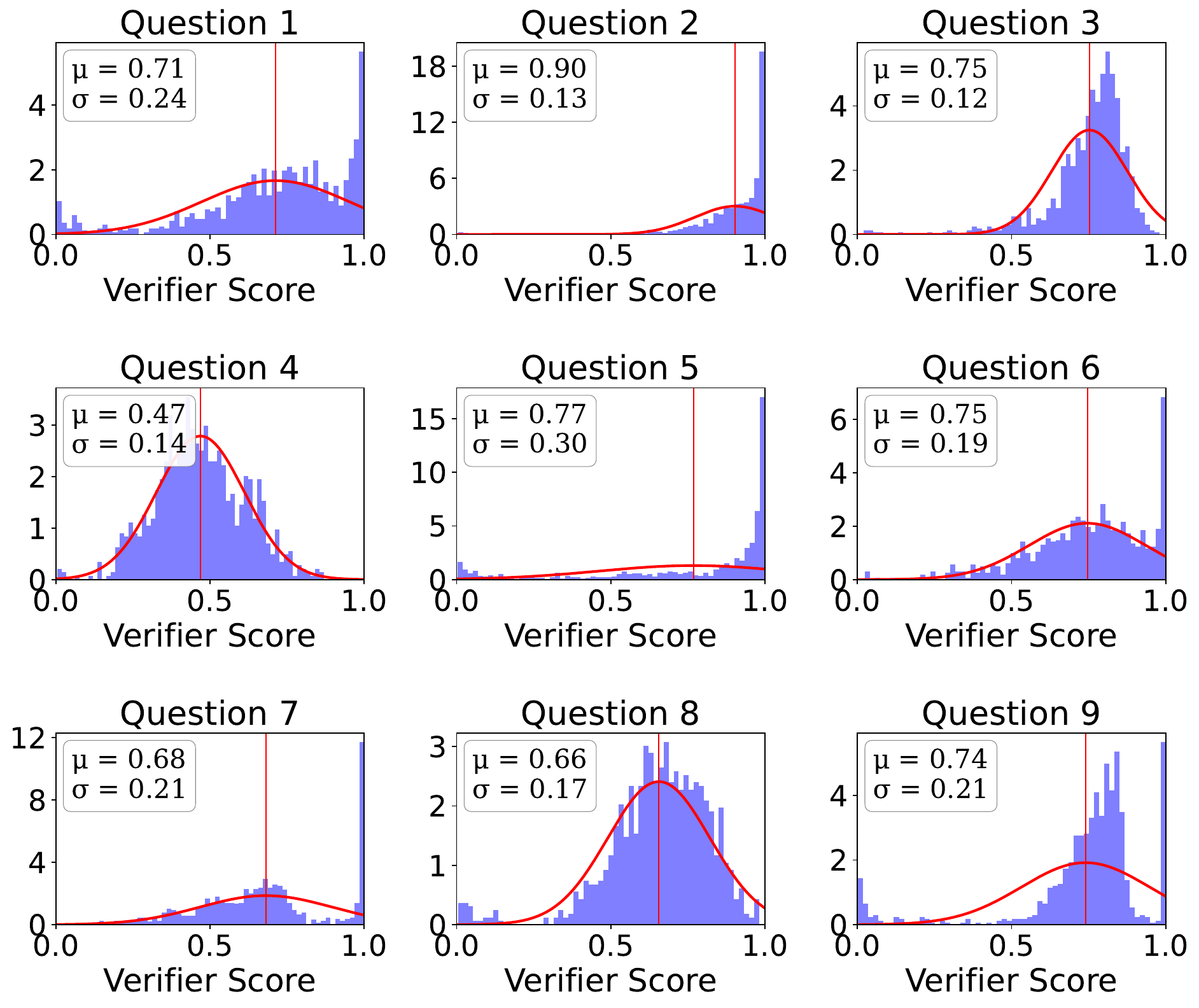}
    \caption{Samples of verifier score distribution: Real vs. Estimated. \textsc{OptScale} accurately fits most distributions under the truncated normal distribution assumption.}
    \label{fig:verifier_score_distribution}
\end{figure}

\subsection{Can \textsc{OptScale} Estimate the Verifier Score Distribution Accurately?}
To evaluate our approach, we compare the estimated distributions against the ground truth. 
Figure \ref{fig:verifier_score_distribution} shows various alignment cases between our modeled distributions and empirical observations.
In most cases (e.g., Questions 1, 2, 3, 4, and 8), the ground truth verifier scores follow a unimodal distribution, which is accurately captured by a truncated normal distribution over $[0,1]$.
For bimodal distributions (e.g., Questions 6 and 7), \textsc{OptScale} maintains reasonably accurate predictions. 
Only in rare instances do we encounter more complex distributions: cases with multiple peaks (Question 9) or extremely narrow, high peaks (Question 5) which challenge the truncated normal approximation. However, \textsc{OptScale} can still capture the main distribution quite well.
These exceptions are also statistically insignificant and do not materially impact overall prediction quality.

\subsection{How is \textsc{OptScale} Sensitive to Its Parameters?}

\textsc{OptScale} employs two hyperparameters, namely the quality threshold $s_{\min}$ and the desired confidence level $\alpha$.
We repeat 66 combinations of these two parameters by varying $s_{\min}$ in the range of [0.80, 0.99] and $\alpha$ in [0.90, 0.99] with a step size of 0.02 and 0.01 respectively to evaluate its sensitivity. We employ Llama-3.2-1B-Instruct backbone with GSM8K benchmark for this focused study. 

\begin{figure}[t!]
    \centering
    \includegraphics[width=0.99\linewidth]{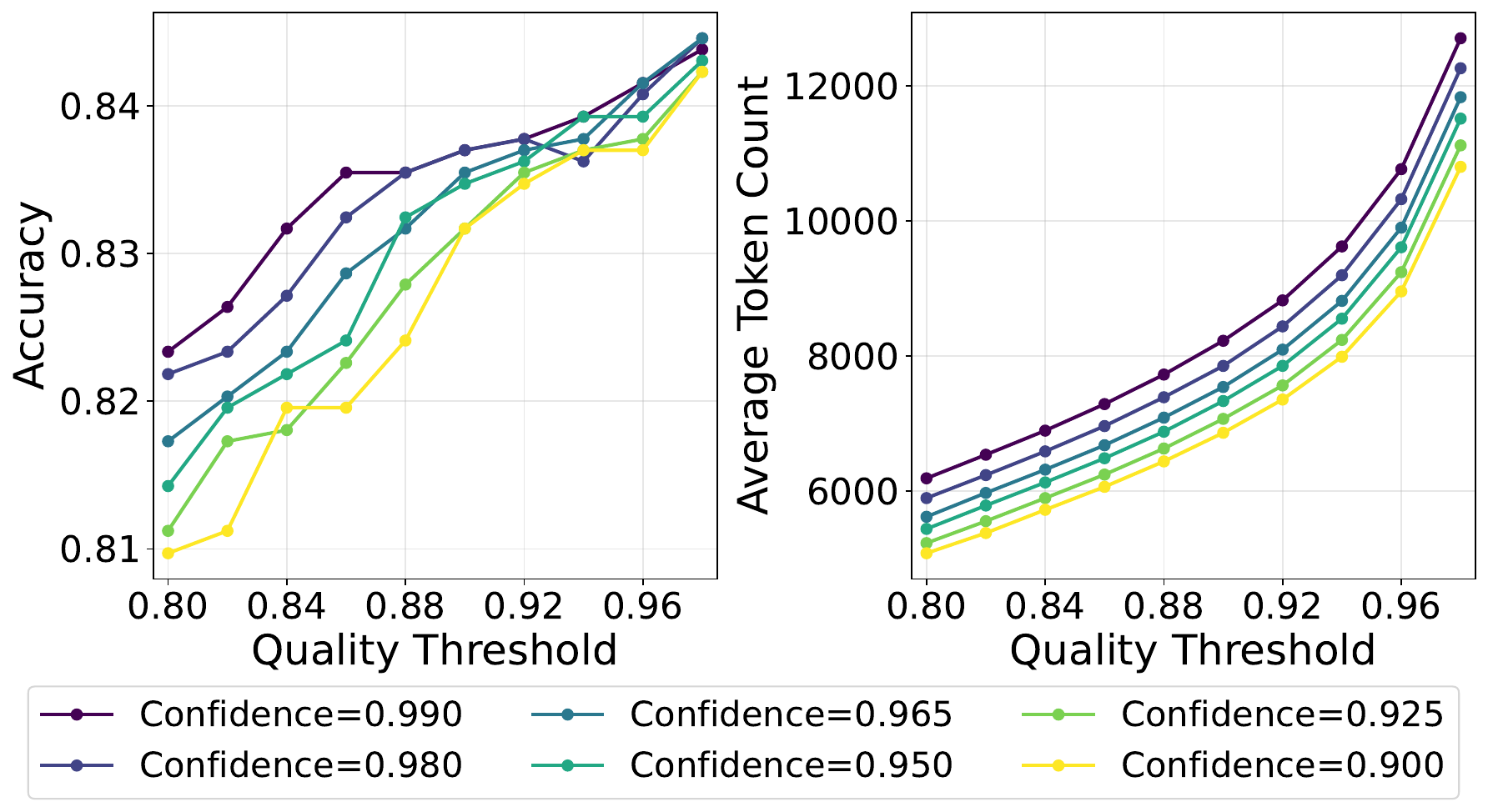}
    \caption{Sensitivity analysis of quality threshold $s_{\min}$: Model performance across target scores.}
    \label{fig:ablation_threshold}
\end{figure}

\textbf{Impact of the Quality Threshold.} Figure \ref{fig:ablation_threshold} shows the changes in accuracy and token consumption corresponding to the quality threshold $s_{\min}$ changes. As $s_{\min}$ increases, the accuracy growth gradually slows down and converges. However, token consumption continues to grow exponentially, with the token consumption multiplying by 2.5 times when the quality threshold increases from 0.8 to 0.99. Therefore, it is recommended to adopt a reasonably high threshold based on available computational budget as the cost-effectiveness of scaling declines sharply when $s_{\min}$ is close to 100\%.

\begin{figure}[t!]
    \centering
    \includegraphics[width=0.99\linewidth]{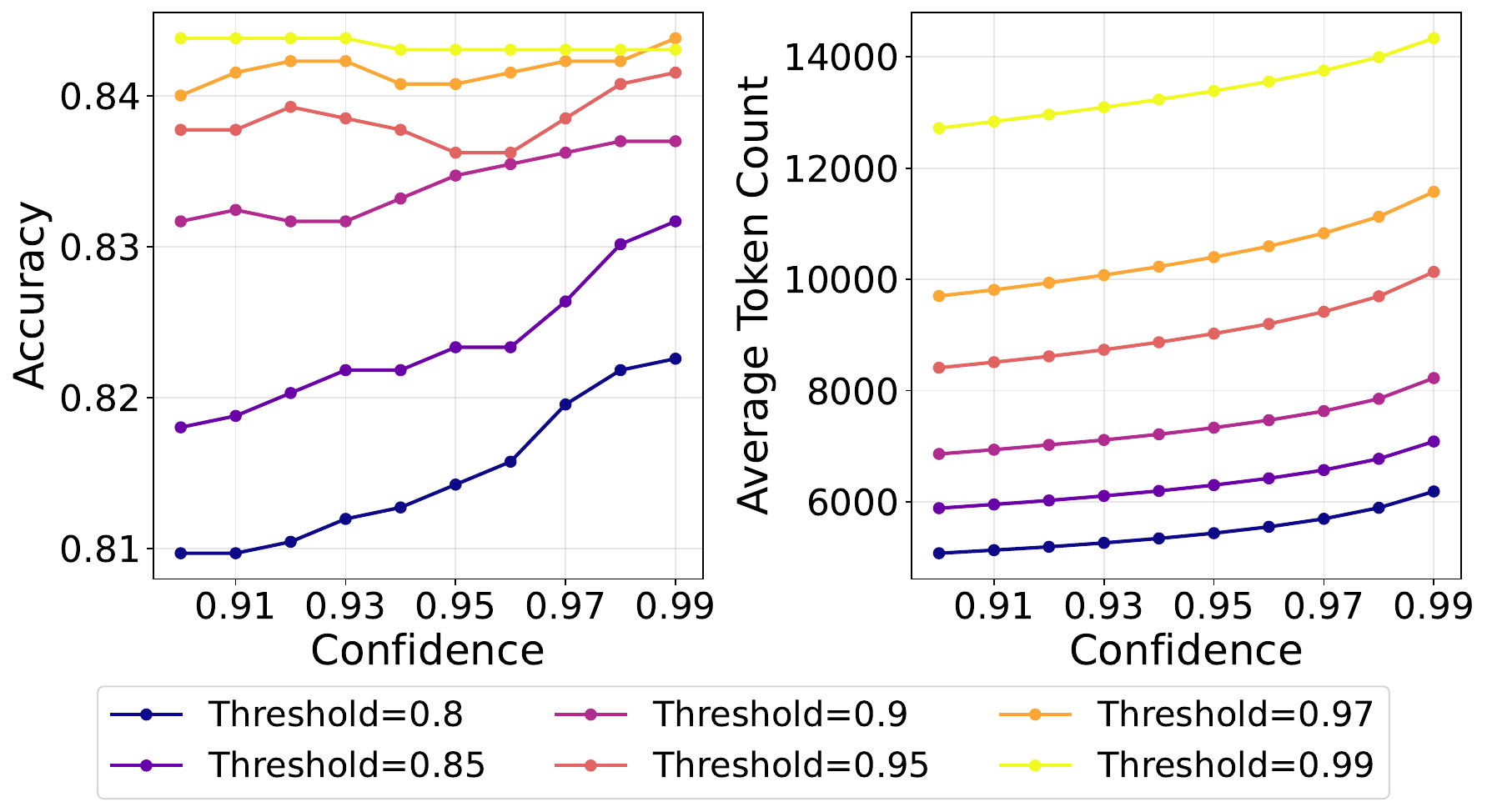}
    \caption{Sensitivity analysis of confidence level $\alpha$: Model performance across target levels.}
    \label{fig:ablation_confidence}
\end{figure}

\textbf{Impact of the Confidence Level $\alpha$.} Similarly, Figure~\ref{fig:ablation_confidence} shows that both accuracy and token consumption increase with confidence level $\alpha$. Unlike the exponential growth in completion tokens observed when increasing $s_{\min}$, token consumption rises nearly linearly with $\alpha$---growing by approximately 100 tokens for every 0.01 increase in confidence. Meanwhile, the overall efficiency experiences only a modest decline as confidence grows.

Overall, we find the quality threshold $s_{\min}$ more sensitive in terms of determining both accuracy and token consumption compared to the confidence level $\alpha$, making its selection much more important. During empirical experiments, we noticed that choosing a very high quality threshold (e.g., 0.99) can harm efficiency and result in suboptimal utilization of completion tokens, while a low quality threshold smaller than 0.9 could lead to accuracy degradation. A quality threshold $s_{\min}$ of around 0.95 to 0.96 is usually recommended, while there's much more room for selecting confidence level $\alpha$: the default could be set to 0.9. 

\subsection{Can \textsc{OptScale} Achieve a New SOTA Performance across Benchmarks?}
\label{subsec:sota}

Table~\ref{tab:sota_results} reports the overall results of different inference-time scaling methods on commonly used mathematical reasoning benchmarks. After careful comparison, we summarize key findings below:

\textbf{State-of-the-art Performance.} When using the same backbone across different benchmarks, \textsc{OptScale} consistently achieves SOTA accuracy while drastically reducing inference token consumption. For example, using the Deepseek-R1-Distill-Qwen-7B backbone with $N=64$, both \textsc{OptScale}$^0$ and \textsc{OptScale}$^t$ achieve the highest accuracy of 94.6\% on MATH-500, significantly surpassing all baselines while concurrently saving 37.0\% (\textsc{OptScale}$^0$) and 56.3\% (\textsc{OptScale}$^t$) than the BoN baseline. 

\textbf{Robustness Across Different Backbones.} The performance of \textsc{OptScale} remains robust and highly efficient across a diverse range of backbone models. On Deepseek-R1-Distill-Qwen-7B, it achieves an average token reduction of 28.9\% (\textsc{OptScale}$^0$) and 32.7\% (\textsc{OptScale}$^t$) among all benchmarks. On Llama3.1-8B-Instruct, the reductions are 23.7\% and 22.9\%, while on QwQ-32B, such reductions even reach 45.0\% and 38.5\% for \textsc{OptScale}$^0$ and \textsc{OptScale}$^t$ respectively. This shows that \textsc{OptScale} has a consistent optimal superiority across different backbones.

Overall, these results demonstrate that \textsc{OptScale} has strong capabilities to optimize the accuracy-efficiency tradeoff across models and benchmarks. The probabilistic foundation of \textsc{OptScale} enables it to dynamically adjust sampling effort based on query difficulty, delivering superior efficiency without compromising reasoning quality.

\subsection{\textsc{OptScale}$^0$ and \textsc{OptScale}$^t$: Which to Choose?} 
\label{subsec:0ort}

Both \textsc{OptScale}$^0$ and \textsc{OptScale}$^t$ exhibit obvious superiority over other baseline methods. However, when it comes to comparing the performance of these two variants of \textsc{OptScale} internally, there's no definite conclusion which one is better. Comprehensively considering both accuracy and token consumption, \textsc{OptScale}$^0$ wins 16 out of 30 sets of comparisons in Table~\ref{tab:sota_results}, while \textsc{OptScale}$^t$ wins 14 times. Despite so, we observe a few prominent features that might guide readers in selecting the more useful variant of \textsc{OptScale} for their task: 

\textbf{1)} \textsc{OptScale}$^t$ seems to favor very difficult benchmarks. On the hardest benchmark AIME25, \textsc{OptScale}$^t$ defeats \textsc{OptScale}$^0$ every time. This is likely because generated answers to difficult questions are more likely to have highly varying and unstable scores. \textsc{OptScale}$^0$ uses MLE to estimate optimal $(\mu^{*}, \sigma^{*})$, but since the initial sampled data is unstable, MLE refinement could be very unstable. \textsc{OptScale}$^t$ first evaluates questions' difficulty using the predictor and then refines the prediction using MAP. This makes it much more stable and more likely to make a well-informed prediction of optimal $N$. 
\textbf{2)} There's a minor tendency that \textsc{OptScale}$^t$ seems to be better at larger $N$ values, while  \textsc{OptScale}$^0$ is better at smaller $N$. This probably indicates that \textsc{OptScale}$^t$ is better at long-term scaling planning, while \textsc{OptScale}$^0$ is better for small-scale scaling given a limited computational budget, especially since it is completely training-free. 

\section{Conclusion}

We introduce a principled probabilistic framework for parallel inference-time scaling in LLM reasoning, addressing the limitations of existing heuristic-based approaches. By modeling the Best-of-$N$ selection under independently and identically distributed (i.i.d.) assumptions, we derive a theoretical lower bound on the sample size required to achieve target performance, offering the first formal guidance for compute-efficient scaling. 
Building on top of this, we propose \textsc{OptScale}, a practical algorithm that dynamically adjusts the number of samples using an LLM-based predictor. 
Extensive experiments across various reasoning benchmarks show that \textsc{OptScale} achieves superior performance with significantly reduced computational cost. 
Our work provides both theoretical foundations and practical implementations for more efficient and adaptive inference-time scaling.
\section{Acknowledgments}

This work is supported in part by the National Natural Science Foundation of China (Grant No. 62372314). The experimental part of this work was supported by The Centre for Large AI Models (CLAIM) of The Hong Kong Polytechnic University. This work is also supported by Hong Kong Research Grants Council under the Theme-based Research Scheme (project no. T43-513/23-N), and is also supported in part by the PolyU Postdoc Matching Fund Scheme (4-W40Z). We also thank the support from PolyU Industrial Centre (IC), PolyU Graduate School (GS), Ren Da, Prof. Li Jing, Prof. Li Qing, and Marco Bettoni.

\bibliography{reference}

@article{zhang2025and,
  title={What, How, Where, and How Well? A Survey on Test-Time Scaling in Large Language Models},
  author={Zhang, Qiyuan and Lyu, Fuyuan and Sun, Zexu and Wang, Lei and Zhang, Weixu and Guo, Zhihan and Wang, Yufei and King, Irwin and Liu, Xue and Ma, Chen},
  journal={arXiv preprint arXiv:2503.24235},
  year={2025}
}

@inproceedings{wang2023self,
  title={Self-Consistency Improves Chain of Thought Reasoning in Language Models},
  author={Wang, Xuezhi and Wei, Jason and Schuurmans, Dale and Le, Quoc V and Chi, Ed H and Narang, Sharan and Chowdhery, Aakanksha and Zhou, Denny},
  booktitle={The Eleventh International Conference on Learning Representations},
  year={2023}
}

@article{tian2025think,
  title={Think Twice: Enhancing LLM Reasoning by Scaling Multi-round Test-time Thinking},
  author={Tian, Xiaoyu and Zhao, Sitong and Wang, Haotian and Chen, Shuaiting and Ji, Yunjie and Peng, Yiping and Zhao, Han and Li, Xiangang},
  journal={arXiv preprint arXiv:2503.19855},
  year={2025}
}

@article{grattafiori2024llama,
  title={The llama 3 herd of models},
  author={Grattafiori, Aaron and Dubey, Abhimanyu and Jauhri, Abhinav and Pandey, Abhinav and Kadian, Abhishek and Al-Dahle, Ahmad and Letman, Aiesha and Mathur, Akhil and Schelten, Alan and Vaughan, Alex and others},
  journal={arXiv preprint arXiv:2407.21783},
  year={2024}
}

@inproceedings{hendrycks2measuring,
  title={Measuring Mathematical Problem Solving With the MATH Dataset},
  author={Hendrycks, Dan and Burns, Collin and Kadavath, Saurav and Arora, Akul and Basart, Steven and Tang, Eric and Song, Dawn and Steinhardt, Jacob},
  booktitle={Thirty-fifth Conference on Neural Information Processing Systems Datasets and Benchmarks Track (Round 2)},
  year={2021}
}

@article{cobbe2021training,
  title={Training verifiers to solve math word problems},
  author={Cobbe, Karl and Kosaraju, Vineet and Bavarian, Mohammad and Chen, Mark and Jun, Heewoo and Kaiser, Lukasz and Plappert, Matthias and Tworek, Jerry and Hilton, Jacob and Nakano, Reiichiro and others},
  journal={arXiv preprint arXiv:2110.14168},
  year={2021}
}

@misc{maa2024aime,
  title = {American Invitational Mathematics Examination - AIME},
  author = {MAA},
  year = {2024},
  month = {February},
  howpublished = {\url{https://huggingface.co/datasets/AI-MO/aimo-validation-aime}}
}

@misc{meta2024llama3,
  title = {Llama 3.2: Revolutionizing edge AI and vision with open, customizable models},
  author = {Meta},
  journal = {Meta},
  year = {2024},
  month = {September},
  howpublished = {\url{https://ai.meta.com/blog/llama-3-2-connect-2024-vision-edge-mobile-devices/}}
}

@article{guo2025deepseek,
  title={{DeepSeek-R1}: Incentivizing Reasoning Capability in LLMs via Reinforcement Learning},
  author={Guo, Daya and Yang, Dejian and Zhang, Haowei and Song, Junxiao and Zhang, Ruoyu and Xu, Runxin and Zhu, Qihao and Ma, Shirong and Wang, Peiyi and Bi, Xiao and others},
  journal={arXiv preprint arXiv:2501.12948},
  year={2025}
}

@article{xu2025towards,
  title={Towards Large Reasoning Models: A Survey of Reinforced Reasoning with Large Language Models},
  author={Xu, Fengli and Hao, Qianyue and Zong, Zefang and Wang, Jingwei and Zhang, Yunke and Wang, Jingyi and Lan, Xiaochong and Gong, Jiahui and Ouyang, Tianjian and Meng, Fanjin and others},
  journal={arXiv preprint arXiv:2501.09686},
  year={2025}
}

@article{yu2024natural,
  title={Natural language reasoning, a survey},
  author={Yu, Fei and Zhang, Hongbo and Tiwari, Prayag and Wang, Benyou},
  journal={ACM Computing Surveys},
  volume={56},
  number={12},
  pages={1--39},
  year={2024},
  publisher={ACM New York, NY}
}

@inproceedings{chongwahngo2008trecvid,
        title={Beyond semantic search: What you observe may not be what you think},
        author={Chong-Wah Ngo and Yu-Gang Jiang and Xiao-Yong Wei and Wanlei Zhao and Feng Wang and Xiao Wu and Hung-Khoon Tan},
        year={2008},
        booktitle={IEEE Computer Society}
        }

@article{wei2022chain,
  title={Chain-of-thought prompting elicits reasoning in large language models},
  author={Wei, Jason and Wang, Xuezhi and Schuurmans, Dale and Bosma, Maarten and Xia, Fei and Chi, Ed and Le, Quoc V and Zhou, Denny and others},
  journal={Advances in neural information processing systems},
  volume={35},
  pages={24824--24837},
  year={2022}
}

@article{ahn2024large,
  title={Large language models for mathematical reasoning: Progresses and challenges},
  author={Ahn, Janice and Verma, Rishu and Lou, Renze and Liu, Di and Zhang, Rui and Yin, Wenpeng},
  journal={arXiv preprint arXiv:2402.00157},
  year={2024}
}

@article{liang2024improving,
  title={Improving llm reasoning through scaling inference computation with collaborative verification},
  author={Liang, Zhenwen and Liu, Ye and Niu, Tong and Zhang, Xiangliang and Zhou, Yingbo and Yavuz, Semih},
  journal={arXiv preprint arXiv:2410.05318},
  year={2024}
}

@article{snell2024scaling,
  title={Scaling llm test-time compute optimally can be more effective than scaling model parameters},
  author={Snell, Charlie and Lee, Jaehoon and Xu, Kelvin and Kumar, Aviral},
  journal={arXiv preprint arXiv:2408.03314},
  year={2024}
}

@article{wan2024dynamic,
  title={Dynamic self-consistency: Leveraging reasoning paths for efficient llm sampling},
  author={Wan, Guangya and Wu, Yuqi and Chen, Jie and Li, Sheng},
  journal={arXiv preprint arXiv:2408.17017},
  year={2024}
}

@article{yao2023tree,
  title={Tree of thoughts: Deliberate problem solving with large language models},
  author={Yao, Shunyu and Yu, Dian and Zhao, Jeffrey and Shafran, Izhak and Griffiths, Tom and Cao, Yuan and Narasimhan, Karthik},
  journal={Advances in neural information processing systems},
  volume={36},
  pages={11809--11822},
  year={2023}
}

@article{long2023large,
  title={Large language model guided tree-of-thought},
  author={Long, Jieyi},
  journal={arXiv preprint arXiv:2305.08291},
  year={2023}
}

@inproceedings{chongwahngo2007trecvid,
        title={Experimenting vireo-374: Bag-of-visual-words and visual-based ontology for semantic video indexing and search},
        author={Chong-Wah Ngo and Yu-Gang Jiang and Xiao-Yong Wei and Feng Wang and Wanlei Zhao and Hung-Khoon Tan and Xiao Wu},
        year={2007},
        booktitle={IEEE Computer Society}
        }

@article{setlur2024rewarding,
  title={Rewarding progress: Scaling automated process verifiers for llm reasoning},
  author={Setlur, Amrith and Nagpal, Chirag and Fisch, Adam and Geng, Xinyang and Eisenstein, Jacob and Agarwal, Rishabh and Agarwal, Alekh and Berant, Jonathan and Kumar, Aviral},
  journal={arXiv preprint arXiv:2410.08146},
  year={2024}
}

@article{prmlessons,
  title={The Lessons of Developing Process Reward Models in Mathematical Reasoning}, 
  author={
    Zhenru Zhang and Chujie Zheng and Yangzhen Wu and Beichen Zhang and Runji Lin and Bowen Yu and Dayiheng Liu and Jingren Zhou and Junyang Lin
  },
  journal={arXiv preprint arXiv:2501.07301},
  year={2025}
}

@article{bai2022training,
  title={Training a helpful and harmless assistant with reinforcement learning from human feedback},
  author={Bai, Yuntao and Jones, Andy and Ndousse, Kamal and Askell, Amanda and Chen, Anna and DasSarma, Nova and Drain, Dawn and Fort, Stanislav and Ganguli, Deep and Henighan, Tom and others},
  journal={arXiv preprint arXiv:2204.05862},
  year={2022}
}

@article{chen2025seal,
  title={SEAL: Steerable Reasoning Calibration of Large Language Models for Free},
  author={Chen, Runjin and Zhang, Zhenyu and Hong, Junyuan and Kundu, Souvik and Wang, Zhangyang},
  journal={arXiv preprint arXiv:2504.07986},
  year={2025}
}

@article{ke2025survey,
  title={A Survey of Frontiers in LLM Reasoning: Inference Scaling, Learning to Reason, and Agentic Systems},
  author={Ke, Zixuan and Jiao, Fangkai and Ming, Yifei and Nguyen, Xuan-Phi and Xu, Austin and Long, Do Xuan and Li, Minzhi and Qin, Chengwei and Wang, Peifeng and Savarese, Silvio and others},
  journal={arXiv preprint arXiv:2504.09037},
  year={2025}
}

@article{ding2025dynamic,
  title={Dynamic parallel tree search for efficient llm reasoning},
  author={Ding, Yifu and Jiang, Wentao and Liu, Shunyu and Jing, Yongcheng and Guo, Jinyang and Wang, Yingjie and Zhang, Jing and Wang, Zengmao and Liu, Ziwei and Du, Bo and others},
  journal={arXiv preprint arXiv:2502.16235},
  year={2025}
}

@article{yang2025towards,
  title={Towards thinking-optimal scaling of test-time compute for llm reasoning},
  author={Yang, Wenkai and Ma, Shuming and Lin, Yankai and Wei, Furu},
  journal={arXiv preprint arXiv:2502.18080},
  year={2025}
}

@article{qu2025optimizing,
  title={Optimizing test-time compute via meta reinforcement fine-tuning},
  author={Qu, Yuxiao and Yang, Matthew YR and Setlur, Amrith and Tunstall, Lewis and Beeching, Edward Emanuel and Salakhutdinov, Ruslan and Kumar, Aviral},
  journal={arXiv preprint arXiv:2503.07572},
  year={2025}
}

@article{yang2025qwen3,
  title={Qwen3 technical report},
  author={Yang, An and Li, Anfeng and Yang, Baosong and Zhang, Beichen and Hui, Binyuan and Zheng, Bo and Yu, Bowen and Gao, Chang and Huang, Chengen and Lv, Chenxu and others},
  journal={arXiv preprint arXiv:2505.09388},
  year={2025}
}

@article{aggarwal2023let,
  title={Let's Sample Step by Step: Adaptive-Consistency for Efficient Reasoning and Coding with LLMs},
  author={Aggarwal, Pranjal and Madaan, Aman and Yang, Yiming and others},
  journal={arXiv preprint arXiv:2305.11860},
  year={2023}
}

@article{li2024escape,
  title={Escape sky-high cost: Early-stopping self-consistency for multi-step reasoning},
  author={Li, Yiwei and Yuan, Peiwen and Feng, Shaoxiong and Pan, Boyuan and Wang, Xinglin and Sun, Bin and Wang, Heda and Li, Kan},
  journal={arXiv preprint arXiv:2401.10480},
  year={2024}
}

@article{wang2024make,
  title={Make every penny count: Difficulty-adaptive self-consistency for cost-efficient reasoning},
  author={Wang, Xinglin and Feng, Shaoxiong and Li, Yiwei and Yuan, Peiwen and Zhang, Yueqi and Tan, Chuyi and Pan, Boyuan and Hu, Yao and Li, Kan},
  journal={arXiv preprint arXiv:2408.13457},
  year={2024}
}

\appendix                  
\section{Appendix}

\subsection{Additional Implementation Details}

We additionally compare our \textsc{OptScale} against the following parallel inference-time scaling methods that adopt early-stopping for token efficiency: (1) \textbf{Adaptive Self-Consistency (ASC)}~\cite{aggarwal2023let}: ASC samples one by one, and stops sampling when the existing sample answers have established a clear majority as judged by a
lightweight stopping criterion. (2) \textbf{Early-stopping Self-Consistency (ESC)}~\cite{li2024escape}: ESC divides the large preset sample size into several sequential small windows, and stops sampling when answers within a window are all the same. (3) \textbf{Difficulty-Adaptive Self-Consistency (DSC)}~\cite{wang2024make}: DSC leverages the difficulty information of
batch queries from both prior and posterior perspectives to adaptively allocate inference resources. 

To ensure a fair comparison across all methods, we predefine four different token levels for each combination of backbone and dataset benchmark. 
This approach aligns token consumption and allows us to evaluate accuracy under a fixed token budget. 
For instance, the four token levels used for Deepseek-R1-Distill-Qwen-7B on the AIME24 dataset benchmark are 50,000, 110,000, 170,000, and 230,000 tokens. 
The token counts for all methods in this group are set around these values, enabling a fair assessment of accuracy within the same budget constraints.

Since we are comparing \textsc{OptScale}$^0$ and \textsc{OptScale}$^t$ with methods specifically designed to minimize token usage, we adopt highly token-efficient hyperparameter settings for both variants. 
Specifically, both methods use a quality threshold $\tau$ of 0.7 for the two AIME datasets, 0.75 for AMC 2023, 0.85 for MATH500, and 0.9 for GSM8K. The confidence level $\alpha$ is consistently set to 0.1 across all datasets.
It is important to note that these hyperparameter settings are not tuned for achieving the highest possible accuracy, but rather to minimize token consumption for a fair comparison with other token-efficient parallel scaling methods. 
In practice, our approach is designed to optimize token usage given accuracy guarantees, rather than focusing solely on reducing token consumption.
We conduct experiments across 6 backbone models and 5 dataset benchmarks, each evaluated at 4 different token levels. This results in a total of 120 distinct setting groups for comparing our methods.

\begin{figure*}[th!]
    \centering
    \includegraphics[width=1\textwidth]{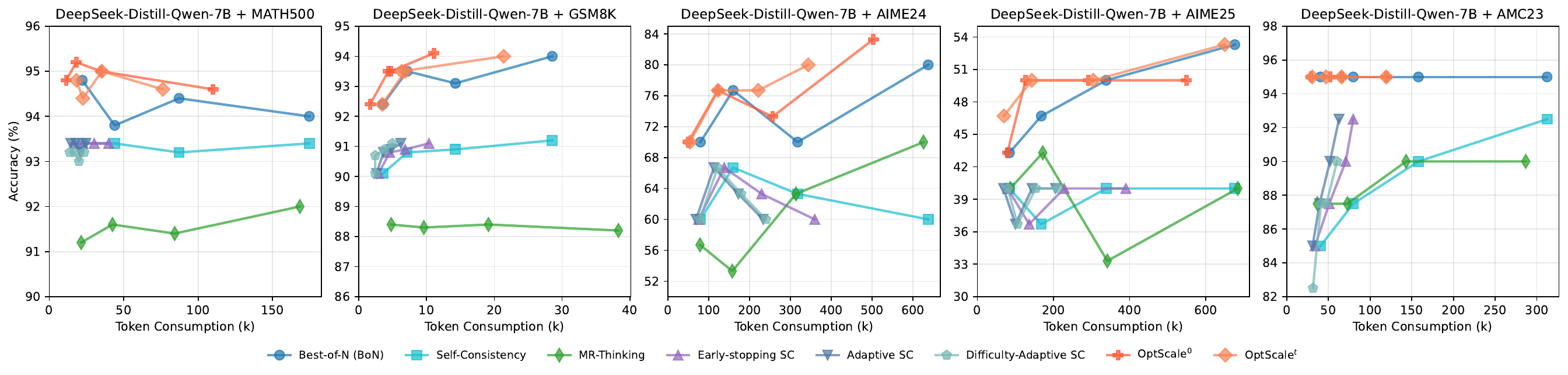}
    \caption{Accuracy-token consumption of different methods when using Deepseek-R1-Distill-Qwen-7B as the backbone model.}
    \label{fig:scale_comp_distill_qwen7B}
\end{figure*}

\begin{figure*}[th!]
    \centering
    \includegraphics[width=1\textwidth]{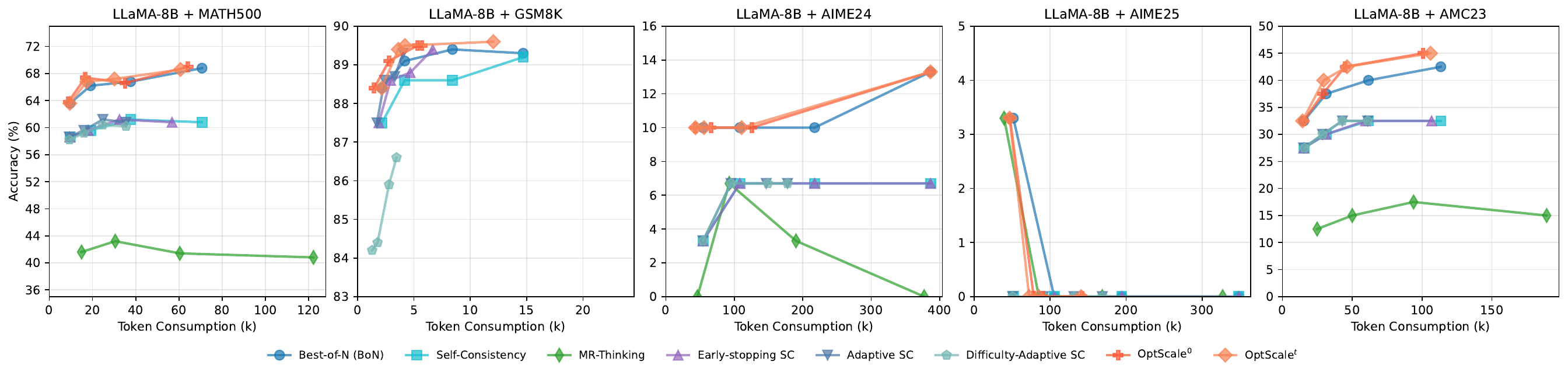}
    \caption{Accuracy-token consumption of different methods when using Llama3.2-8B-Instruct as the backbone model.}
    \label{fig:scale_comp_llama8B}
\end{figure*}

\begin{figure*}[th!]
    \centering
    \includegraphics[width=1\textwidth]{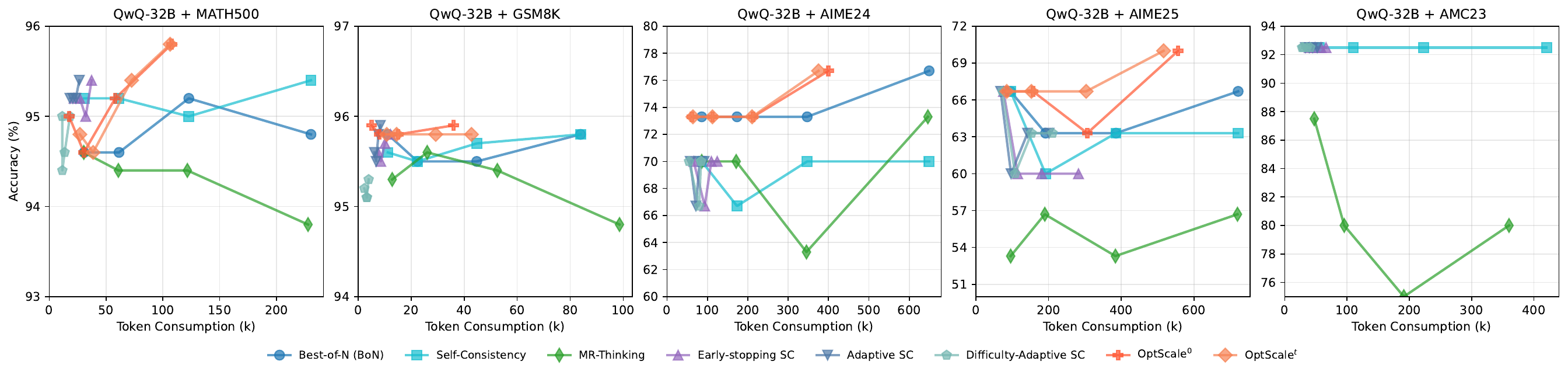}
    \caption{Accuracy-token consumption of different methods when using QwQ-32B as the backbone model.}
    \label{fig:scale_comp_QwQ32B}
\end{figure*}

\subsection{Full Experimental Results}

\begin{figure*}[ht!]
    \centering
    \includegraphics[width=1\textwidth]{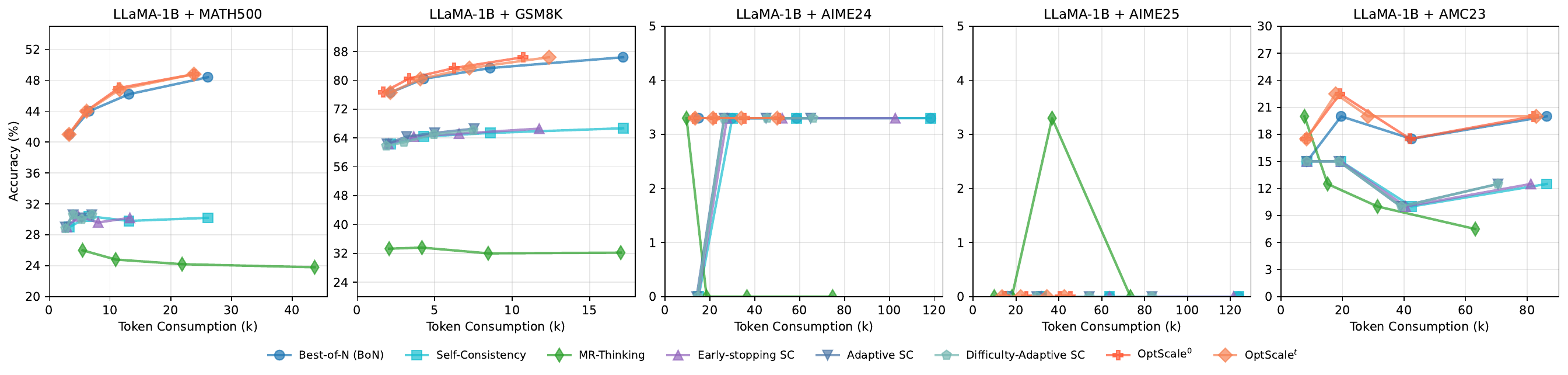}
    \caption{Accuracy-token consumption of different methods when using Llama3.1-1B-Instruct as the backbone model.}
    \label{fig:scale_comp_llama1B}
\end{figure*}

Figures~\ref{fig:scale_comp_distill_qwen7B} $\sim$ \ref{fig:scale_comp_llama1B} present the accuracy - token comparisons as we scale up compute using each method during inference-time.
Notably, all \textsc{OptScale} implementations along the frontier consistently outperform baselines, delivering higher accuracy with lower computational cost (token consumption). This confirms its ability to maintain scaling optimality across configurations.

The final results of different inference-time scaling methods are presented in Tables~\ref{tab:Distill-Qwen-7B} $\sim$ \ref{tab:Llama3.2-1B-Instruct}. We highlight the following key findings:
(1) \textbf{State-of-the-art Performance (Accuracy and Token Efficiency).} When using the same backbone across different benchmarks, \textsc{OptScale} consistently achieves state-of-the-art (SOTA) accuracy while drastically reducing inference token consumption. Even though we give \textsc{OptScale}$^0$ and \textsc{OptScale}$^t$ extremely limited token budgets, which are even lower than the token consumption of all token-saving algorithms, including ASC, ESC, and DSC, \textsc{OptScale} still significantly outperforms all baselines in accuracy. This demonstrates the effectiveness of OptScale's token optimization mechanism, which is based on probabilistic guarantees of accuracy.
(2) \textbf{Robustness to Different Backbones.} The performance of \textsc{OptScale} remains robust and highly efficient across a diverse range of backbone models. For example, on Deepseek-R1-Distill-Qwen-7B, it achieves an average token reduction of 28.9\% (\textsc{OptScale}$^0$) and 32.7\% (\textsc{OptScale}$^t$) among all benchmarks. This shows that \textsc{OptScale} has a consistent optimal superiority across different backbones.

These results demonstrate strong capabilities of \textsc{OptScale} to optimize the accuracy-efficiency tradeoff across diverse models and benchmarks. The method's probabilistic foundation enables it to dynamically adjust sampling effort based on query difficulty, delivering SOTA efficiency without compromising solution quality.

\subsection{Discussions on Robustness and Generalization}

This section presents additional analyses that extend the main paper, with a focus on robustness to sampling correlation, distributional approximation, and verifier variation, as well as generalization to broader evaluation domains.

\paragraph{Robustness to Sampling Correlation.} 

Parallel samples can exhibit varying degrees of dependence depending on the decoding temperature. To assess the robustness of \textsc{OptScale} under such conditions, we evaluated its performance across a range of sampling temperatures from 0.2 to 1.0, which induce different levels of correlation among generated completions.
Table~\ref{tab:iid-table} presents results on AIME24 using Deepseek-R1-Distill-Qwen-7B as the backbone model. Under our default setting, \textsc{OptScale} improves accuracy by 1.7\% while reducing token usage by 22.0\% compared to Best-of-N.
When averaged over all temperature settings, the gains are even more significant: 2.5\% accuracy improvement and 35.3\% token reduction, highlighting the robustness of \textsc{OptScale} under varying degrees of sampling correlation.

\begin{figure*}[ht!]
    \centering
    \includegraphics[width=1\textwidth]{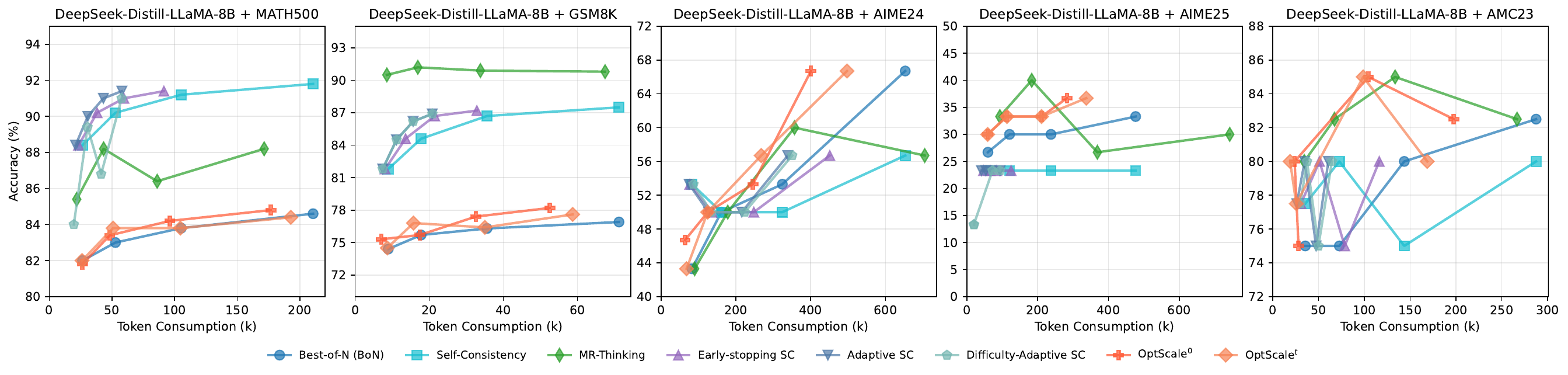}
    \caption{Accuracy-token consumption of different methods when using Deepseek-R1-Distill-Llama-8B as the backbone model.}
    \label{fig:scale_comp_distill_llama8B}
\end{figure*}

\begin{figure*}[ht!]
    \centering
    \includegraphics[width=1\textwidth]{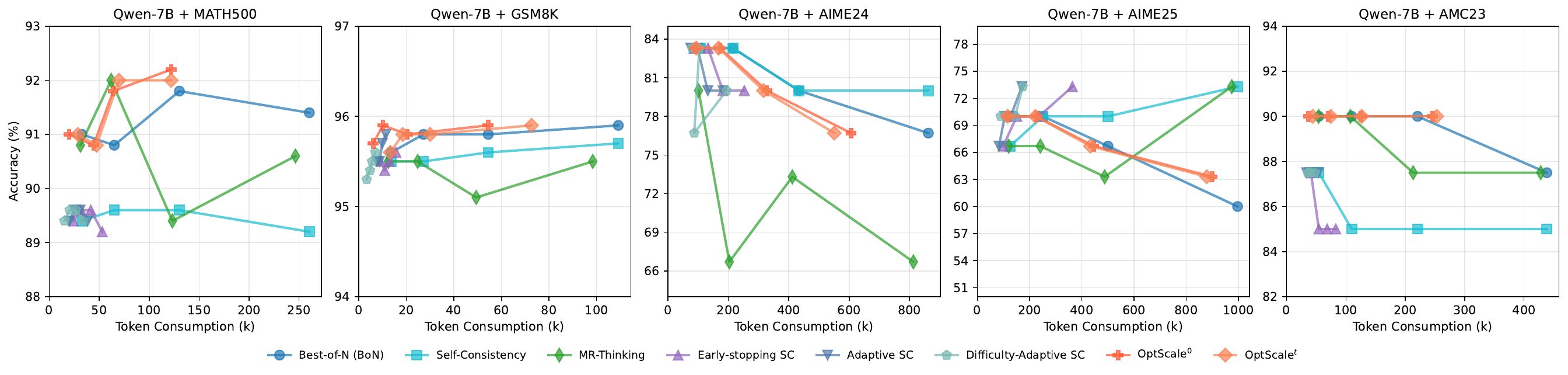}
    \caption{Accuracy-token consumption of different methods when using Qwen3-8B as the backbone model.}
    \label{fig:scale_comp_qwen37B}
\end{figure*}

\begin{table*}[t!]
\centering
\resizebox{0.96\textwidth}{!}{%
\setlength{\tabcolsep}{4pt}
\begin{tabular}{l l c c c c c c c c c c}
\toprule
\multirow{2}{*}{\textbf{Level}} & \multirow{2}{*}{\textbf{Baseline Method}} & \multicolumn{2}{c}{\textbf{MATH-500}} & \multicolumn{2}{c}{\textbf{GSM8K}} & \multicolumn{2}{c}{\textbf{AIME 2024}} & \multicolumn{2}{c}{\textbf{AIME 2025}} & \multicolumn{2}{c}{\textbf{AMC 2023}} \\
\cmidrule{3-12}
 & & \textbf{Acc.}  & \textbf{Toks.} ($\downarrow$) & \textbf{Acc.} & \textbf{Toks.} ($\downarrow$) & \textbf{Acc.}  & \textbf{Toks.} ($\downarrow$) & \textbf{Acc.} & \textbf{Toks.} ($\downarrow$) & \textbf{Acc.}  & \textbf{Toks.} ($\downarrow$) \\
\midrule
\multicolumn{12}{l}{\textbf{Deepseek-R1-Distill-Qwen-7B}} \\
\midrule
\multirow{8}{*}{Level 1} & Best-of-N (BoN) & 94.6 & 16557 & 92.3 & 3142 & 66.7 & 58680 & \textbf{43.3} & 63346 & \textbf{95.0} & 30356 \\
 & Self-Consistency & 93.8 & 16557 & 89.7 & 3142 & 56.7 & 58680 & 36.7 & 63346 & 87.5 & 30356 \\
 & MR-Thinking & 91.8 & 15966 & 88.8 & 2968 & 56.7 & 58240 & 30.0 & 64709 & 92.5 & 32287 \\
 & Early-stopping SC (ESC) & 93.2 & 15679 & 89.7 & 2720 & 56.7 & 57293 & 36.7 & 61767 & 87.5 & 31138 \\
 & Adaptive SC (ASC) & 93.8 & 15526 & 90.3 & 2792 & 56.7 & 53714 & 36.7 & 63372 & 85.0 & 31029 \\
 & Difficulty-Adaptive SC (DSC) & 93.2 & 15208 & 90.4 & 2969 & 56.7 & 62885 & 36.7 & 63801 & 82.5 & 31541 \\
 \cmidrule{2-12}
 & \textsc{OptScale}$^0$ & \textbf{95.0} & 14927 & \textbf{93.2} & \textbf{1908} & \textbf{70.0} & \textbf{49505} & 40.0 & \textbf{58073} & 92.5 & \textbf{24711} \\
 & \textsc{OptScale}$^t$ & 94.6 & \textbf{14474} & 91.9 & 2649 & \textbf{70.0} & 49993 & \textbf{43.3} & 58632 & \textbf{95.0} & 28530 \\
\midrule
\multirow{8}{*}{Level 2} & Best-of-N (BoN) & 94.8 & 19352 & 92.4 & 3582 & \textbf{70.0} & 120032 & 40.0 & 116266 & \textbf{95.0} & 40511 \\
 & Self-Consistency & 93.2 & 19352 & 90.1 & 3582 & 56.7 & 120032 & 36.7 & 116266 & 85.0 & 40511 \\
 & MR-Thinking & 91.0 & 18670 & 88.9 & 3568 & 53.3 & 118944 & \textbf{50.0} & 118987 & 87.5 & 41434 \\
 & Early-stopping SC (ESC) & 93.6 & 17547 & 90.4 & 3422 & 63.3 & 115299 & 36.7 & 110510 & 85.0 & 41540 \\
 & Adaptive SC (ASC) & 93.6 & 17207 & 90.9 & 3330 & 66.7 & 112390 & 36.7 & 112243 & 87.5 & 40378 \\
 & Difficulty-Adaptive SC (DSC) & 93.2 & 17095 & 90.7 & 3447 & 63.3 & 111681 & 40.0 & 112765 & 87.5 & 41095 \\
  \cmidrule{2-12}
 & \textsc{OptScale}$^0$ & \textbf{95.0} & 16924 & \textbf{93.2} & \textbf{1908} & \textbf{70.0} & 107003 & 43.3 & \textbf{106495} & 92.5 & \textbf{24711} \\
 & \textsc{OptScale}$^t$ & 94.8 & \textbf{16427} & 92.3 & 3074 & \textbf{70.0} & \textbf{99734} & 46.7 & 106571 & \textbf{95.0} & 39633 \\
\midrule
\multirow{8}{*}{Level 3} & Best-of-N (BoN) & 94.8 & 22135 & 92.6 & 4022 & \textbf{73.3} & 170348 & \textbf{46.7} & 168080 & \textbf{95.0} & 50597 \\
 & Self-Consistency & 93.4 & 22135 & 90.3 & 4022 & 63.3 & 170348 & 36.7 & 168080 & 85.0 & 50597 \\
 & MR-Thinking & 91.2 & 21396 & 88.5 & 4185 & 60.0 & 177489 & 30.0 & 161489 & 90.0 & 54303 \\
 & Early-stopping SC (ESC) & 93.4 & 20560 & 90.9 & 4004 & 63.3 & 173184 & 40.0 & 165704 & 87.5 & 51009 \\
 & Adaptive SC (ASC) & 93.2 & 20105 & 91.0 & 4024 & 63.3 & 172043 & 40.0 & 160147 & 90.0 & 50606 \\
 & Difficulty-Adaptive SC (DSC) & 93.2 & 20016 & 90.8 & 4668 & 63.3 & 173404 & 40.0 & 162381 & 90.0 & 51690 \\
  \cmidrule{2-12}
 & \textsc{OptScale}$^0$ & \textbf{95.2} & \textbf{18170} & \textbf{93.2} & \textbf{1908} & 70.0 & 107003 & 43.3 & \textbf{154222} & 92.5 & \textbf{24711} \\
 & \textsc{OptScale}$^t$ & 94.8 & 19930 & 92.6 & 3909 & 70.0 & \textbf{99734} & \textbf{46.7} & 155590 & \textbf{95.0} & 41812 \\
\midrule
\multirow{8}{*}{Level 4} & Best-of-N (BoN) & 94.8 & 24836 & 93.1 & 5353 & \textbf{73.3} & 231270 & \textbf{46.7} & 200530 & \textbf{95.0} & 60655 \\
 & Self-Consistency & 93.6 & 24836 & 90.4 & 5353 & 63.3 & 231270 & 36.7 & 200530 & 85.0 & 60655 \\
 & MR-Thinking & 91.2 & 24045 & 88.4 & 5380 & 60.0 & 235263 & 36.7 & 204021 & 87.5 & 63660 \\
 & Early-stopping SC (ESC) & 93.4 & 23145 & 90.9 & 5052 & 63.3 & 234303 & 40.0 & 201347 & 90.0 & 60255 \\
 & Adaptive SC (ASC) & 93.4 & 23076 & 90.8 & 5021 & 60.0 & 231145 & 40.0 & 200614 & 92.5 & 60015 \\
 & Difficulty-Adaptive SC (DSC) & 93.2 & 23002 & 90.8 & 5026 & 60.0 & 232995 & 40.0 & 211646 & 92.5 & 61433 \\
  \cmidrule{2-12}
 & \textsc{OptScale}$^0$ & \textbf{95.2} & \textbf{18170} & \textbf{93.2} & \textbf{1908} & 70.0 & 107003 & 43.3 & 160260 & 92.5 & \textbf{24711} \\
 & \textsc{OptScale}$^t$ & 94.6 & 21525 & 92.9 & 4748 & 70.0 & \textbf{99734} & \textbf{46.7} & \textbf{155590} & \textbf{95.0} & 41812 \\
\bottomrule
\end{tabular}
}
\caption{Comparison of different inference-time scaling methods on Deepseek-R1-Distill-Qwen-7B across various reasoning benchmarks. ``Level'' denotes the token consumption level, ``Acc.'' denotes accuracy (\%), ``Toks.'' indicate the total number of inference tokens.}
\label{tab:Distill-Qwen-7B}
\end{table*}
\begin{table*}[t!]
\centering
\resizebox{0.96\textwidth}{!}{%
\setlength{\tabcolsep}{4pt}
\begin{tabular}{l l c c c c c c c c c c}
\toprule
\multirow{2}{*}{\textbf{Level}} & \multirow{2}{*}{\textbf{Baseline Method}} & \multicolumn{2}{c}{\textbf{MATH-500}} & \multicolumn{2}{c}{\textbf{GSM8K}} & \multicolumn{2}{c}{\textbf{AIME 2024}} & \multicolumn{2}{c}{\textbf{AIME 2025}} & \multicolumn{2}{c}{\textbf{AMC 2023}} \\
\cmidrule{3-12}
 & & \textbf{Acc.}  & \textbf{Toks.} ($\downarrow$) & \textbf{Acc.} & \textbf{Toks.} ($\downarrow$) & \textbf{Acc.}  & \textbf{Toks.} ($\downarrow$) & \textbf{Acc.} & \textbf{Toks.} ($\downarrow$) & \textbf{Acc.}  & \textbf{Toks.} ($\downarrow$) \\
\midrule
\multicolumn{12}{l}{\textbf{Llama3.1-8B-Instruct}} \\
\midrule
\multirow{8}{*}{Level 1} & Best-of-N (BoN) & \textbf{63.6} & 9609 & \textbf{88.0} & 1892 & \textbf{10.0} & 54212 & \textbf{3.3} & 51510 & \textbf{32.5} & 15321 \\
 & Self-Consistency & 58.6 & 9609 & 87.2 & 1892 & 3.3 & 54212 & 0.0 & 51510 & 27.5 & 15321 \\
 & MR-Thinking & 40.8 & 11047 & 63.3 & 2092 & 3.3 & 60115 & 0.0 & 53518 & 20.0 & 16778 \\
 & Early-stopping SC (ESC) & 58.6 & 9449 & 87.5 & 1878 & 3.3 & 54212 & 0.0 & 51510 & 27.5 & 15292 \\
 & Adaptive SC (ASC) & 58.6 & 9245 & 87.6 & 1833 & 3.3 & 54212 & 0.0 & 50586 & 27.5 & 15234 \\
 & Difficulty-Adaptive SC (DSC) & 58.2 & 9321 & 87.6 & 1837 & 3.3 & 54212 & 0.0 & 50586 & 27.5 & 15831 \\
\cmidrule{2-12}
 & \textsc{OptScale}$^0$ & 62.4 & 8760 & \textbf{88.0} & \textbf{1177} & 6.7 & \textbf{51499} & \textbf{3.3} & 49981 & 22.5 & 14794 \\
 & \textsc{OptScale}$^t$ & 63.0 & \textbf{8134} & 87.5 & 1628 & \textbf{10.0} & 53896 & \textbf{3.3} & \textbf{49603} & \textbf{32.5} & \textbf{14655} \\
\midrule
\multirow{8}{*}{Level 2} & Best-of-N (BoN) & \textbf{66.2} & 16602 & 88.3 & 2427 & \textbf{10.0} & 92920 & 0.0 & 94592 & \textbf{35.0} & 29034 \\
 & Self-Consistency & 60.2 & 16602 & 87.6 & 2427 & 6.7 & 92920 & 0.0 & 94592 & 30.0 & 29034 \\
 & MR-Thinking & 40.4 & 17046 & 62.1 & 2471 & 6.7 & 92594 & 0.0 & 91647 & 10.0 & 28715 \\
 & Early-stopping SC (ESC) & 59.0 & 16798 & \textbf{88.4} & 2439 & 6.7 & 92920 & 0.0 & 94592 & 30.0 & 28855 \\
 & Adaptive SC (ASC) & 60.0 & 16716 & \textbf{88.4} & 2444 & 6.7 & 95051 & 0.0 & 91174 & 30.0 & 28698 \\
 & Difficulty-Adaptive SC (DSC) & 59.2 & 16009 & \textbf{88.4} & 2455 & 6.7 & 95051 & 0.0 & 91174 & 30.0 & 28122 \\
 \cmidrule{2-12}
 & \textsc{OptScale}$^0$ & 62.4 & \textbf{8760} & 88.0 & \textbf{1177} & 6.7 & 83189 & 0.0 & 89018 & 22.5 & \textbf{23052} \\
 & \textsc{OptScale}$^t$ & 63.8 & 10494 & 88.3 & 2398 & \textbf{10.0} & \textbf{70345} & 0.0 & \textbf{84000} & 27.5 & 27464 \\
\midrule
\multirow{8}{*}{Level 3} & Best-of-N (BoN) & \textbf{66.2} & 25200 & 88.6 & 2904 & \textbf{10.0} & 141501 & 0.0 & 131351 & \textbf{40.0} & 41570 \\
 & Self-Consistency & 61.2 & 25200 & \textbf{88.9} & 2904 & 6.7 & 141501 & 0.0 & 131351 & 27.5 & 41570 \\
 & MR-Thinking & 40.8 & 24781 & 63.5 & 3144 & 0.0 & 148153 & \textbf{3.3} & 130290 & 22.5 & 42513 \\
 & Early-stopping SC (ESC) & 61.8 & 24501 & 88.6 & 2928 & 6.7 & 141501 & 0.0 & 131351 & 27.5 & 40889 \\
 & Adaptive SC (ASC) & 61.2 & 24188 & 88.7 & 2917 & 6.7 & 142630 & 0.0 & 131323 & 30.0 & 40783 \\
 & Difficulty-Adaptive SC (DSC) & 60.6 & 24222 & 88.6 & 3069 & 6.7 & 142630 & 0.0 & 131323 & 30.0 & 40575 \\
 & \textsc{OptScale}$^0$ & 62.4 & \textbf{8760} & 88.0 & \textbf{1177} & 6.7 & 83189 & 0.0 & 94514 & 22.5 & \textbf{23052} \\
 \cmidrule{2-12}
 & \textsc{OptScale}$^t$ & 63.8 & 10494 & 88.3 & 2398 & \textbf{10.0} & \textbf{70345} & 0.0 & \textbf{84000} & 27.5 & 29653 \\
\midrule
\multirow{8}{*}{Level 4} & Best-of-N (BoN) & \textbf{66.6} & 35264 & \textbf{89.0} & 3670 & \textbf{10.0} & 176183 & 0.0 & 164505 & \textbf{40.0} & 60174 \\
 & Self-Consistency & 61.0 & 35264 & 88.6 & 3670 & 6.7 & 176183 & 0.0 & 164505 & 30.0 & 60174 \\
 & MR-Thinking & 40.0 & 36256 & 60.7 & 3524 & 0.0 & 179415 & 0.0 & 163915 & 12.5 & 63077 \\
 & Early-stopping SC (ESC) & 61.2 & 35128 & 88.4 & 3461 & 6.7 & 176183 & 0.0 & 164505 & 32.5 & 60824 \\
 & Adaptive SC (ASC) & 61.0 & 35180 & 88.8 & 3432 & 3.3 & 175174 & 0.0 & 160075 & 32.5 & 60508 \\
 & Difficulty-Adaptive SC (DSC) & 60.4 & 35231 & 88.1 & 3448 & 3.3 & 175174 & 0.0 & 160075 & 32.5 & 61202 \\
 \cmidrule{2-12}
 & \textsc{OptScale}$^0$ & 62.4 & \textbf{8760} & 88.0 & \textbf{1177} & 6.7 & 83189 & 0.0 & 94514 & 22.5 & \textbf{23052} \\
 & \textsc{OptScale}$^t$ & 63.8 & 10494 & 88.3 & 2398 & \textbf{10.0} & \textbf{70345} & 0.0 & \textbf{84000} & 27.5 & 29653 \\
\bottomrule
\end{tabular}
}
\caption{Comparison of different inference-time scaling methods on Llama3.1-8B-Instruct across various reasoning benchmarks. ``Level'' denotes the token consumption level, ``Acc.'' denotes accuracy (\%), ``Toks.'' indicate the total number of inference tokens.}
\label{tab:Llama-3.1-8B-Instruct}
\end{table*}

\begin{table*}[t!]
\centering
\resizebox{0.96\textwidth}{!}{%
\setlength{\tabcolsep}{4pt}
\begin{tabular}{l l c c c c c c c c c c}
\toprule
\multirow{2}{*}{\textbf{Level}} & \multirow{2}{*}{\textbf{Baseline Method}} & \multicolumn{2}{c}{\textbf{MATH-500}} & \multicolumn{2}{c}{\textbf{GSM8K}} & \multicolumn{2}{c}{\textbf{AIME 2024}} & \multicolumn{2}{c}{\textbf{AIME 2025}} & \multicolumn{2}{c}{\textbf{AMC 2023}} \\ 
\cmidrule{3-12}
 & & \textbf{Acc.}  & \textbf{Toks.} ($\downarrow$) & \textbf{Acc.} & \textbf{Toks.} ($\downarrow$) & \textbf{Acc.}  & \textbf{Toks.} ($\downarrow$) & \textbf{Acc.} & \textbf{Toks.} ($\downarrow$) & \textbf{Acc.}  & \textbf{Toks.} ($\downarrow$) \\
\midrule
\multicolumn{12}{l}{\textbf{Qwen/QwQ-32B}} \\ 
\midrule
\multirow{8}{*}{Level 1} & Best-of-N (BoN) & 95.0 & 15204 & \textbf{95.3} & 2800 & 70.0 & 64564 & \textbf{70.0} & 83231 & 92.5 & 34171 \\
 & Self-Consistency & \textbf{95.2} & 15204 & 94.8 & 2800 & 70.0 & 64564 & 60.0 & 83231 & 92.5 & 34171 \\
 & MR-Thinking & 94.8 & 15185 & \textbf{95.3} & 2889 & \textbf{73.3} & 63782 & 56.7 & 83252 & 80.0 & 35249 \\
& Early-stopping SC (ESC) & \textbf{95.2} & 15204 & 94.8 & 2800 & 66.7 & 62102 & 63.3 & 81550 & 92.5 & 34171 \\
& Adaptive SC (ASC) & \textbf{95.2} & 15204 & 94.8 & 2800 & 66.7 & 60047 & 63.3 & 82850 & 92.5 & 33854 \\
& Difficulty-Adaptive SC (DSC) & \textbf{95.2} & 15204 & 94.8 & 2800 & 66.7 & 60055 & 63.3 & 81482 & 92.5 & 33926 \\
 \cmidrule{2-12}
& \textsc{OptScale}$^0$ & 94.8 & \textbf{11459} & 94.8 & \textbf{1412} & \textbf{73.3} & \textbf{56453} & \textbf{70.0} & 76131 & 92.5 & \textbf{33074} \\
& \textsc{OptScale}$^t$ & 94.8 & \textbf{11459} & 94.8 & \textbf{1412} & \textbf{73.3} & 58638 & \textbf{70.0} & \textbf{75281} & \textbf{95.0} & 33076 \\
\midrule
\multirow{8}{*}{Level 2} & Best-of-N (BoN) & 95.0 & 15204 & \textbf{95.3} & 2800 & 73.3 & 75359 & \textbf{66.7} & 131275 & \textbf{95.0} & 40608 \\
 & Self-Consistency & 95.2 & 15204 & 94.8 & 2800 & 66.7 & 75359 & 60.0 & 131275 & 92.5 & 40608 \\
& MR-Thinking & 94.8 & 15185 & \textbf{95.3} & 2889 & 60.0 & 75185 & 63.3 & 131330 & 80.0 & 41301 \\
& Early-stopping SC (ESC) & 95.2 & 15204 & 94.8 & 2800 & 66.7 & 70176 & 56.7 & 122917 & 92.5 & 37442 \\
& Adaptive SC (ASC) & 95.2 & 15204 & 94.8 & 2800 & 66.7 & 70407 & 63.3 & 120413 & 92.5 & 36030 \\
& Difficulty-Adaptive SC (DSC) & 95.2 & 15204 & 94.8 & 2800 & 66.7 & 77270 & 60.0 & 120920 & 92.5 & 36794 \\
 \cmidrule{2-12}
& \textsc{OptScale}$^0$ & \textbf{95.4} & 13994 & 94.8 & \textbf{1412} & \textbf{76.7} & \textbf{66564} & \textbf{66.7} & 113438 & 92.5 & 34979 \\
& \textsc{OptScale}$^t$ & 94.8 & \textbf{11459} & 94.8 & \textbf{1412} & \textbf{76.7} & 69738 & \textbf{66.7} & \textbf{111539} & \textbf{95.0} & \textbf{33076} \\
\midrule
\multirow{8}{*}{Level 3} & Best-of-N (BoN) & 95.0 & 19074 & 95.5 & 4184 & 73.3 & 85853 & 63.3 & 167636 & \textbf{95.0} & 40608 \\
& Self-Consistency & 94.8 & 19074 & \textbf{95.7} & 4184 & 70.0 & 85853 & 63.3 & 167636 & 92.5 & 40608 \\
& MR-Thinking & 94.0 & 19013 & 95.5 & 4549 & 70.0 & 85864 & 63.3 & 165932 & 80.0 & 41301 \\
& Early-stopping SC (ESC) & 94.8 & 19074 & \textbf{95.7} & 4184 & 66.7 & 81401 & 56.7 & 160581 & 92.5 & 39228 \\
& Adaptive SC (ASC) & 94.8 & 16001 & \textbf{95.7} & 4184 & 70.0 & 80512 & 60.0 & 161016 & 92.5 & 38324 \\
& Difficulty-Adaptive SC (DSC) & 94.8 & 16495 & \textbf{95.7} & 4184 & 70.0 & 81798 & 63.3 & 161049 & 92.5 & 38186 \\
 \cmidrule{2-12}
& \textsc{OptScale}$^0$ & \textbf{95.2} & 15209 & 95.4 & \textbf{2713} & \textbf{76.7} & \textbf{76638} & 63.3 & 158828 & 92.5 & 37251 \\
& \textsc{OptScale}$^t$ & 95.0 & \textbf{14636} & 95.3 & 2800 & \textbf{76.7} & 79813 & \textbf{66.7} & \textbf{153375} & \textbf{95.0} & \textbf{37104} \\
\midrule
\multirow{8}{*}{Level 4} & Best-of-N (BoN) & 95.0 & 19074 & 95.5 & 4184 & 73.3 & 85853 & 63.3 & 203712 & \textbf{95.0} & 40608 \\
& Self-Consistency & 94.8 & 19074 & \textbf{95.7} & 4184 & 70.0 & 85853 & 60.0 & 203712 & 92.5 & 40608 \\
& MR-Thinking & 94.0 & 19013 & 95.5 & 4549 & 70.0 & 85864 & 53.3 & 201341 & 80.0 & 41301 \\
& Early-stopping SC (ESC) & 94.8 & 19074 & \textbf{95.7} & 4184 & 66.7 & 85082 & 60.0 & 203264 & 92.5 & 40642 \\
& Adaptive SC (ASC) & \textbf{95.2} & 18165 & \textbf{95.7} & 4184 & 70.0 & 84320 & 63.3 & 201114 & 92.5 & 40124 \\
& Difficulty-Adaptive SC (DSC) & 95.0 & 18477 & \textbf{95.7} & 4184 & 70.0 & 91185 & 63.3 & 200369 & 92.5 & 40154 \\
 \cmidrule{2-12}
& \textsc{OptScale}$^0$ & 95.0 & \textbf{17469} & 95.4 & \textbf{2713} & 73.3 & 81913 & 63.3 & \textbf{195085} & 92.5 & 39471 \\
& \textsc{OptScale}$^t$ & \textbf{95.2} & 17831 & 95.3 & 2800 & \textbf{76.7} & \textbf{79813} & \textbf{70.0} & 199249 & \textbf{95.0} & \textbf{37104} \\
\bottomrule
\end{tabular}
}
\caption{Comparison of different inference-time scaling methods on QwQ-32B across various reasoning benchmarks. ``Level'' denotes the token consumption level, ``Acc.'' denotes accuracy (\%), ``Toks.'' indicate the total number of inference tokens.}
\label{tab:QwQ-32B}
\end{table*}

\begin{table*}[t!]
\centering
\resizebox{0.96\textwidth}{!}{%
\setlength{\tabcolsep}{4pt}
\begin{tabular}{l l c c c c c c c c c c}
\toprule
\multirow{2}{*}{\textbf{Level}} & \multirow{2}{*}{\textbf{Baseline Method}} & \multicolumn{2}{c}{\textbf{MATH-500}} & \multicolumn{2}{c}{\textbf{GSM8K}} & \multicolumn{2}{c}{\textbf{AIME 2024}} & \multicolumn{2}{c}{\textbf{AIME 2025}} & \multicolumn{2}{c}{\textbf{AMC 2023}} \\
\cmidrule{3-12}
 & & \textbf{Acc.}  & \textbf{Toks.} ($\downarrow$) & \textbf{Acc.} & \textbf{Toks.} ($\downarrow$) & \textbf{Acc.}  & \textbf{Toks.} ($\downarrow$) & \textbf{Acc.} & \textbf{Toks.} ($\downarrow$) & \textbf{Acc.}  & \textbf{Toks.} ($\downarrow$) \\
\midrule
\multicolumn{12}{l}{\textbf{Deepseek-R1-Distill-Llama-8B}} \\
\midrule
\multirow{8}{*}{Level 1} & Best-of-N (BoN) & 82.0 & 26497 & 74.8 & 7820 & 43.3 & 82784 & 30.0 & 52161 & 75.0 & 35598 \\
 & Self-Consistency & 88.4 & 26497 & 82.1 & 7820 & \textbf{53.3} & 82784 & 23.3 & 52161 & 77.5 & 35598 \\
 & MR-Thinking & 87.4 & 27355 & \textbf{91.1} & 7501 & 46.7 & 77994 & \textbf{36.7} & 47399 & 80.0 & 34500 \\
 & Early-stopping SC (ESC) & 88.8 & 25668 & 81.8 & 8072 & \textbf{53.3} & 77010 & 23.3 & 47092 & 80.0 & 36737 \\
 & Adaptive SC (ASC) & \textbf{89.4} & 26123 & 81.8 & 7427 & \textbf{53.3} & 75140 & 26.7 & 45364 & 80.0 & 34899 \\
 & Difficulty-Adaptive SC (DSC) & 88.6 & 27328 & 81.8 & 7437 & \textbf{53.3} & 86706 & 26.7 & 48102 & 77.5 & 34325 \\
  \cmidrule{2-12}
 & \textsc{OptScale}$^0$ & 80.4 & \textbf{16742} & 74.7 & \textbf{4808} & 46.7 & 74160 & 23.3 & \textbf{42207} & 92.5 & \textbf{33074} \\
 & \textsc{OptScale}$^t$ & 81.0 & 23163 & 74.8 & 6569 & 43.3 & \textbf{73513} & 23.3 & 42671 & \textbf{95.0} & 33076 \\
\midrule
\multirow{8}{*}{Level 2} & Best-of-N (BoN) & 82.4 & 33329 & 75.0 & 11183 & 43.3 & 131376 & 26.7 & 59548 & 72.5 & 40194 \\
 & Self-Consistency & 89.2 & 33329 & 83.5 & 11183 & \textbf{53.3} & 131376 & 23.3 & 59548 & 82.5 & 40194 \\
 & MR-Thinking & 88.0 & 30016 & \textbf{90.9} & 11668 & \textbf{53.3} & 133804 & \textbf{33.3} & 58831 & 77.5 & 42637 \\
 & Early-stopping SC (ESC) & 89.4 & 31237 & 84.2 & 11711 & 50.0 & 136123 & 26.7 & 56115 & 77.5 & 41591 \\
 & Adaptive SC (ASC) & \textbf{90.0} & 30707 & 84.5 & 11155 & 50.0 & 132653 & 23.3 & 55777 & 80.0 & 40419 \\
 & Difficulty-Adaptive SC (DSC) & 89.4 & 30491 & 84.5 & 11183 & 46.7 & 131963 & 20.0 & 55383 & 80.0 & 40411 \\
  \cmidrule{2-12}
 & \textsc{OptScale}$^0$ & 80.4 & \textbf{16742} & 74.7 & \textbf{4808} & 43.3 & \textbf{112742} & 23.3 & \textbf{51987} & 92.5 & 39471 \\
 & \textsc{OptScale}$^t$ & 80.8 & 25751 & 75.1 & 10755 & 40.0 & 113596 & 23.3 & 54625 & \textbf{95.0} & \textbf{37104} \\
\midrule
\multirow{8}{*}{Level 3} & Best-of-N (BoN) & 83.6 & 42903 & 75.5 & 15604 & 46.7 & 224361 & 26.7 & 75205 & 72.5 & 53737 \\
 & Self-Consistency & 89.4 & 42903 & 84.0 & 15604 & 50.0 & 224361 & 26.7 & 75205 & 77.5 & 53737 \\
 & MR-Thinking & 86.2 & 40652 & \textbf{90.7} & 15879 & \textbf{66.7} & 223722 & \textbf{33.3} & 81503 & 77.5 & 50870 \\
 & Early-stopping SC (ESC) & 89.8 & 41367 & 85.7 & 15283 & 50.0 & 221395 & 23.3 & 70507 & 80.0 & 51119 \\
 & Adaptive SC (ASC) & \textbf{91.0} & 40090 & 86.0 & 15202 & 46.7 & 220680 & 26.7 & 70354 & 75.0 & 50420 \\
 & Difficulty-Adaptive SC (DSC) & 90.2 & 41285 & 86.2 & 15673 & 46.7 & 221978 & 20.0 & 72401 & 75.0 & 50423 \\
  \cmidrule{2-12}
 & \textsc{OptScale}$^0$ & 80.4 & \textbf{16742} & 74.7 & \textbf{4808} & 43.3 & \textbf{112742} & 23.3 & \textbf{67524} & \textbf{95.0} & 48249 \\
 & \textsc{OptScale}$^t$ & 80.8 & 25751 & 75.3 & 11815 & 40.0 & 113596 & 23.3 & 69404 & \textbf{95.0} & \textbf{47791} \\
\midrule
\multirow{8}{*}{Level 4} & Best-of-N (BoN) & 83.0 & 56133 & 76.3 & 21094 & 53.3 & 335404 & 26.7 & 90655 & 75.0 & 63179 \\
 & Self-Consistency & 90.0 & 56133 & 85.7 & 21094 & 46.7 & 335404 & 23.3 & 90655 & 80.0 & 63179 \\
 & MR-Thinking & 85.6 & 56777 & \textbf{90.3} & 20096 & \textbf{63.3} & 333869 & \textbf{33.3} & 93307 & 90.0 & 63294 \\
 & Early-stopping SC (ESC) & 90.6 & 56173 & 86.7 & 20161 & 53.3 & 331027 & 26.7 & 91343 & 80.0 & 61799 \\
 & Adaptive SC (ASC) & \textbf{91.0} & 55280 & 86.7 & 20043 & 56.7 & 331362 & 23.3 & 90138 & 80.0 & 60084 \\
 & Difficulty-Adaptive SC (DSC) & 89.2 & 55094 & 86.7 & 20063 & 53.3 & 332892 & 23.3 & 90965 & 77.5 & 60154 \\
  \cmidrule{2-12}
 & \textsc{OptScale}$^0$ & 80.4 & \textbf{16742} & 74.7 & \textbf{4808} & 43.3 & \textbf{112742} & 23.3 & 84280 & 95.0 & \textbf{57707} \\
 & \textsc{OptScale}$^t$ & 80.8 & 25751 & 75.3 & 11815 & 40.0 & 113596 & 23.3 & \textbf{75228} & \textbf{97.5} & 57761 \\
\bottomrule
\end{tabular}
}
\caption{Comparison of different inference-time scaling methods on Deepseek-R1-Distill-Llama-8B across various reasoning benchmarks. ``Level'' denotes the token consumption level, ``Acc.'' denotes accuracy (\%), ``Toks.'' indicate the total number of inference tokens.}
\label{tab:Distill-Llama-8B}
\end{table*}

\begin{table*}[t!]
\centering
\resizebox{0.96\textwidth}{!}{%
\setlength{\tabcolsep}{4pt}
\begin{tabular}{l l c c c c c c c c c c}
\toprule
\multirow{2}{*}{\textbf{Level}} & \multirow{2}{*}{\textbf{Baseline Method}} & \multicolumn{2}{c}{\textbf{MATH-500}} & \multicolumn{2}{c}{\textbf{GSM8K}} & \multicolumn{2}{c}{\textbf{AIME 2024}} & \multicolumn{2}{c}{\textbf{AIME 2025}} & \multicolumn{2}{c}{\textbf{AMC 2023}} \\
\cmidrule{3-12}
 & & \textbf{Acc.}  & \textbf{Toks.} ($\downarrow$) & \textbf{Acc.} & \textbf{Toks.} ($\downarrow$) & \textbf{Acc.}  & \textbf{Toks.} ($\downarrow$) & \textbf{Acc.} & \textbf{Toks.} ($\downarrow$) & \textbf{Acc.}  & \textbf{Toks.} ($\downarrow$) \\
\midrule
\multicolumn{12}{l}{\textbf{Qwen3-8B}} \\
\midrule
\multirow{8}{*}{Level 1} & Best-of-N (BoN) & 90.4 & 24433 & 95.6 & 6840 & \textbf{83.3} & 79251 & \textbf{70.0} & 92928 & \textbf{90.0} & 41049 \\
 & Self-Consistency & 89.4 & 24433 & 95.4 & 6840 & \textbf{83.3} & 79251 & 66.7 & 92928 & 87.5 & 41049 \\
 & MR-Thinking & 88.0 & 23379 & 95.2 & 6287 & 70.0 & 75734 & 56.7 & 104859 & 87.5 & 40374 \\
 & Early-stopping SC (ESC) & 89.4 & 23860 & 95.4 & 6840 & 80.0 & 79403 & \textbf{70.0} & 91846 & 87.5 & 36439 \\
 & Adaptive SC (ASC) & 90.6 & 23455 & 95.4 & 6840 & \textbf{83.3} & 75299 & \textbf{70.0} & 92068 & 87.5 & 36762 \\
 & Difficulty-Adaptive SC (DSC) & 89.8 & 23363 & 95.4 & 6840 & \textbf{83.3} & 75887 & \textbf{70.0} & 109443 & 87.5 & 37677 \\
  \cmidrule{2-12}
 & \textsc{OptScale}$^0$ & \textbf{91.0} & 22587 & \textbf{95.8} & \textbf{5107} & \textbf{83.3} & \textbf{69691} & \textbf{70.0} & \textbf{87742} & \textbf{90.0} & \textbf{33496} \\
 & \textsc{OptScale}$^t$ & 90.4 & \textbf{22339} & \textbf{95.8} & \textbf{5107} & \textbf{83.3} & 72436 & \textbf{70.0} & 88045 & \textbf{90.0} & 35678 \\
\midrule
\multirow{8}{*}{Level 2} & Best-of-N (BoN) & 90.2 & 28526 & 95.6 & 6840 & \textbf{83.3} & 79251 & \textbf{70.0} & 125791 & \textbf{90.0} & 41049 \\
 & Self-Consistency & 89.6 & 28526 & 95.4 & 6840 & \textbf{83.3} & 79251 & 66.7 & 125791 & 87.5 & 41049 \\
 & MR-Thinking & 90.2 & 27267 & 95.2 & 6287 & 73.3 & 88220 & 66.7 & 120220 & 87.5 & 40374 \\
 & Early-stopping SC (ESC) & \textbf{90.6} & 27162 & 95.4 & 6840 & 80.0 & 79403 & \textbf{70.0} & 125849 & 87.5 & 38645 \\
 & Adaptive SC (ASC) & 90.2 & 27262 & 95.4 & 6840 & 80.0 & 79310 & \textbf{70.0} & 120981 & 87.5 & 37730 \\
 & Difficulty-Adaptive SC (DSC) & 89.6 & 27412 & 95.4 & 6840 & \textbf{83.3} & 79864 & \textbf{70.0} & 120556 & 87.5 & 37677 \\
  \cmidrule{2-12}
 & \textsc{OptScale}$^0$ & 90.4 & 26658 & 95.5 & 5599 & \textbf{83.3} & \textbf{69691} & \textbf{70.0} & \textbf{116109} & \textbf{90.0} & 36618 \\
 & \textsc{OptScale}$^t$ & 90.2 & \textbf{25548} & \textbf{95.8} & \textbf{5107} & \textbf{83.3} & 72436 & \textbf{70.0} & 116413 & \textbf{90.0} & \textbf{35678} \\
\midrule
\multirow{8}{*}{Level 3} & Best-of-N (BoN) & \textbf{91.0} & 32553 & 95.6 & 6840 & \textbf{83.3} & 93305 & 63.3 & 156827 & \textbf{90.0} & 41049 \\
 & Self-Consistency & 89.4 & 32553 & 95.4 & 6840 & 80.0 & 93305 & 70.0 & 156827 & 87.5 & 41049 \\
 & MR-Thinking & 90.8 & 31136 & 95.3 & 7841 & 73.3 & 88220 & 60.0 & 150608 & 87.5 & 40374 \\
 & Early-stopping SC (ESC) & 89.6 & 31656 & 95.4 & 6840 & \textbf{83.3} & 86308 & 70.0 & 151447 & 87.5 & 40944 \\
 & Adaptive SC (ASC) & 89.6 & 31120 & 95.4 & 6840 & 80.0 & 83648 & \textbf{73.3} & 150814 & 87.5 & 39212 \\
 & Difficulty-Adaptive SC (DSC) & 89.4 & 31294 & 95.4 & 6840 & 80.0 & 90409 & 70.0 & 150326 & 87.5 & 40296 \\
  \cmidrule{2-12}
 & \textsc{OptScale}$^0$ & 90.6 & 30648 & 95.7 & 6278 & \textbf{83.3} & \textbf{69691} & 63.3 & 142239 & \textbf{90.0} & 36618 \\
 & \textsc{OptScale}$^t$ & \textbf{91.0} & \textbf{28613} & \textbf{95.8} & \textbf{5107} & \textbf{83.3} & 72436 & 63.3 & \textbf{141162} & \textbf{90.0} & \textbf{35678} \\
\midrule
\multirow{8}{*}{Level 4} & Best-of-N (BoN) & \textbf{91.4} & 36604 & 95.5 & 8536 & \textbf{83.3} & 93305 & 63.3 & 172401 & \textbf{90.0} & 41049 \\
 & Self-Consistency & 90.2 & 36604 & 95.3 & 8536 & 80.0 & 93305 & 70.0 & 172401 & 87.5 & 41049 \\
 & MR-Thinking & 90.2 & 34989 & 95.3 & 7841 & 73.3 & 88220 & 66.7 & 181618 & 87.5 & 40374 \\
 & Early-stopping SC (ESC) & 90.0 & 34783 & 95.3 & 8536 & \textbf{83.3} & 86308 & \textbf{73.3} & 174376 & 87.5 & 40944 \\
 & Adaptive SC (ASC) & 89.4 & 34024 & 95.3 & 7079 & 80.0 & 86519 & \textbf{73.3} & 170936 & 87.5 & 40074 \\
 & Difficulty-Adaptive SC (DSC) & 89.4 & 34037 & 95.6 & 7008 & 80.0 & 90409 & \textbf{73.3} & 173205 & 87.5 & 40296 \\
\cmidrule{2-12}
 & \textsc{OptScale}$^0$ & 90.6 & 33274 & \textbf{95.7} & \textbf{6278} & \textbf{83.3} & \textbf{80071} & 63.3 & 168077 & \textbf{90.0} & \textbf{39763} \\
 & \textsc{OptScale}$^t$ & \textbf{91.4} & \textbf{31590} & 95.6 & 6773 & \textbf{83.3} & 83455 & 63.3 & \textbf{163166} & \textbf{90.0} & 39939 \\
\bottomrule
\end{tabular}
}
\caption{Comparison of different inference-time scaling methods on Qwen3-8B across various reasoning benchmarks. ``Level'' denotes the token consumption level, ``Acc.'' denotes accuracy (\%), ``Toks.'' indicate the total number of inference tokens.}
\label{tab:Qwen3-8B}
\end{table*}
\begin{table*}[t!]
\centering
\resizebox{0.96\textwidth}{!}{%
\setlength{\tabcolsep}{4pt}
\begin{tabular}{l l c c c c c c c c c c}
\toprule
\multirow{2}{*}{\textbf{Level}} & \multirow{2}{*}{\textbf{Baseline Method}} & \multicolumn{2}{c}{\textbf{MATH-500}} & \multicolumn{2}{c}{\textbf{GSM8K}} & \multicolumn{2}{c}{\textbf{AIME 2024}} & \multicolumn{2}{c}{\textbf{AIME 2025}} & \multicolumn{2}{c}{\textbf{AMC 2023}} \\
\cmidrule{3-12}
 & & \textbf{Acc.}  & \textbf{Toks.} ($\downarrow$) & \textbf{Acc.} & \textbf{Toks.} ($\downarrow$) & \textbf{Acc.}  & \textbf{Toks.} ($\downarrow$) & \textbf{Acc.} & \textbf{Toks.} ($\downarrow$) & \textbf{Acc.}  & \textbf{Toks.} ($\downarrow$) \\
\midrule
\multicolumn{12}{l}{\textbf{Llama3.2-1B-Instruct}} \\
\midrule
\multirow{8}{*}{Level 1} & Best-of-N (BoN) & \textbf{41.0} & 3280 & 76.6 & 2143 & \textbf{3.3} & 14916 & 0.0 & 15849 & 15.0 & 8396 \\
 & Self-Consistency & 29.0 & 3280 & 62.4 & 2143 & 0.0 & 14916 & 0.0 & 15849 & 15.0 & 8396 \\
 & MR-Thinking & 24.2 & 3440 & 33.3 & 2058 & \textbf{3.3} & 15274 & 0.0 & 15025 & 12.5 & 8616 \\
 & Early-stopping SC (ESC) & 28.4 & 3145 & 62.4 & 2004 & 0.0 & 14703 & 0.0 & 15849 & 15.0 & 8379 \\
 & Adaptive SC (ASC) & 28.6 & 3011 & 63.5 & 2102 & 0.0 & 14034 & 0.0 & 15849 & 15.0 & 8347 \\
 & Difficulty-Adaptive SC (DSC) & 28.6 & 3037 & 63.0 & 2084 & 0.0 & 14542 & 0.0 & 15849 & 15.0 & 8347 \\
  \cmidrule{2-12}
 & \textsc{OptScale}$^0$ & 40.4 & \textbf{2841} & \textbf{94.8} & \textbf{1412} & \textbf{3.3} & \textbf{12845} & 0.0 & \textbf{14710} & \textbf{17.5} & \textbf{8099} \\
 & \textsc{OptScale}$^t$ & 40.2 & 2846 & \textbf{94.8} & \textbf{1412} & \textbf{3.3} & 13334 & 0.0 & 14984 & \textbf{17.5} & 8290 \\
\midrule
\multirow{8}{*}{Level 2} & Best-of-N (BoN) & 41.8 & 4096 & 78.8 & 3213 & \textbf{3.3} & 26678 & 0.0 & 30274 & \textbf{20.0} & 19528 \\
 & Self-Consistency & 28.6 & 4096 & 64.4 & 3213 & \textbf{3.3} & 26678 & 0.0 & 30274 & 15.0 & 19528 \\
 & MR-Thinking & 24.2 & 4110 & 33.2 & 3107 & 0.0 & 26362 & 0.0 & 31136 & 12.5 & 19079 \\
 & Early-stopping SC (ESC) & 29.4 & 4127 & 64.4 & 3048 & \textbf{3.3} & 26430 & 0.0 & 30274 & 15.0 & 19429 \\
 & Adaptive SC (ASC) & 30.6 & 4017 & 64.3 & 3040 & \textbf{3.3} & 26301 & 0.0 & 31638 & 15.0 & 20333 \\
 & Difficulty-Adaptive SC (DSC) & 29.8 & 4030 & 63.8 & 3014 & \textbf{3.3} & 26301 & 0.0 & 31638 & 15.0 & 20660 \\
 \cmidrule{2-12}
 & \textsc{OptScale}$^0$ & 41.2 & \textbf{3132} & \textbf{95.4} & \textbf{2713} & \textbf{3.3} & \textbf{24939} & 0.0 & 29747 & 17.5 & 18671 \\
 & \textsc{OptScale}$^t$ & \textbf{42.0} & 3648 & 95.3 & 2800 & \textbf{3.3} & 25934 & 0.0 & \textbf{29636} & 15.0 & \textbf{18044} \\
\midrule
\multirow{8}{*}{Level 3} & Best-of-N (BoN) & \textbf{43.2} & 5328 & 81.0 & 5089 & \textbf{3.3} & 45368 & 0.0 & 54746 & \textbf{17.5} & 39611 \\
 & Self-Consistency & 29.4 & 5328 & 64.6 & 5089 & \textbf{3.3} & 45368 & 0.0 & 54746 & 12.5 & 39611 \\
 & MR-Thinking & 26.0 & 5475 & 31.5 & 5243 & \textbf{3.3} & 45543 & 0.0 & 54957 & 5.0 & 39217 \\
 & Early-stopping SC (ESC) & 30.4 & 5036 & 65.2 & 5148 & \textbf{3.3} & 45794 & 0.0 & 54746 & 10.0 & 39597 \\
 & Adaptive SC (ASC) & 31.0 & 5039 & 65.0 & 5088 & \textbf{3.3} & 46490 & 0.0 & 54170 & 10.0 & 39801 \\
 & Difficulty-Adaptive SC (DSC) & 30.6 & 5132 & 64.5 & 5049 & \textbf{3.3} & 45030 & 0.0 & 54170 & 10.0 & 39123 \\
 \cmidrule{2-12}
 & \textsc{OptScale}$^0$ & 41.2 & \textbf{3132} & \textbf{95.5} & \textbf{4011} & \textbf{3.3} & 30932 & 0.0 & 32000 & \textbf{17.5} & \textbf{24216} \\
 & \textsc{OptScale}$^t$ & 42.0 & 3648 & \textbf{95.5} & 4184 & \textbf{3.3} & \textbf{27839} & 0.0 & \textbf{29636} & 15.0 & 26480 \\
\midrule
\multirow{8}{*}{Level 4} & Best-of-N (BoN) & \textbf{44.4} & 6967 & 82.6 & 7241 & \textbf{3.3} & 64646 & 0.0 & 74229 & \textbf{20.0} & 60744 \\
 & Self-Consistency & 30.4 & 6967 & 65.0 & 7241 & \textbf{3.3} & 64646 & 0.0 & 74229 & 12.5 & 60744 \\
 & MR-Thinking & 23.2 & 6896 & 32.9 & 7126 & 0.0 & 64960 & 0.0 & 73375 & 10.0 & 60093 \\
 & Early-stopping SC (ESC) & 30.8 & 6898 & 65.2 & 7078 & \textbf{3.3} & 65706 & 0.0 & 74229 & 12.5 & 60005 \\
 & Adaptive SC (ASC) & 30.6 & 6818 & 66.1 & 7049 & \textbf{3.3} & 64284 & 0.0 & 73470 & 12.5 & 60859 \\
 & Difficulty-Adaptive SC (DSC) & 30.4 & 6818 & 65.9 & 7046 & \textbf{3.3} & 64459 & 0.0 & 73470 & 12.5 & 60859 \\
 \cmidrule{2-12}
 & \textsc{OptScale}$^0$ & 41.2 & \textbf{3132} & \textbf{95.7} & \textbf{6618} & \textbf{3.3} & 30932 & 0.0 & 32000 & 17.5 & \textbf{24216} \\
 & \textsc{OptScale}$^t$ & 42.0 & 3648 & 95.6 & 6960 & \textbf{3.3} & \textbf{27839} & 0.0 & \textbf{29636} & 15.0 & 26480 \\
\bottomrule
\end{tabular}
}
\caption{Comparison of different inference-time scaling methods on Llama3.2-1B-Instruct across various reasoning benchmarks. ``Level'' denotes the token consumption level, ``Acc.'' denotes accuracy (\%), ``Toks.'' indicate the total number of inference tokens.}
\label{tab:Llama3.2-1B-Instruct}
\end{table*}

\begin{table*}[t!]
\centering
\resizebox{0.96\textwidth}{!}{%
\begin{tabular}{l c c c c c c c c c c}
\toprule
\multirow{2}{*}{\textbf{Baseline Method}} & \multicolumn{2}{c}{\textbf{MATH-500}} & \multicolumn{2}{c}{\textbf{GSM8K}} & \multicolumn{2}{c}{\textbf{AIME 2024}} & \multicolumn{2}{c}{\textbf{AIME 2025}} & \multicolumn{2}{c}{\textbf{AMC 2023}} \\ 
\cmidrule{2-11}
 & \textbf{Acc.}  & \textbf{Toks.} ($\downarrow$) & \textbf{Acc.} & \textbf{Toks.} ($\downarrow$) & \textbf{Acc.}  & \textbf{Toks.} ($\downarrow$)  & \textbf{Acc.} & \textbf{Toks.} ($\downarrow$) & \textbf{Acc.}  & \textbf{Toks.} ($\downarrow$) \\
\midrule
\multicolumn{11}{l}{\textbf{Deepseek-R1-Distill-Qwen-7B}} \\ 
\midrule
Best-of-N (BoN) ($N=8$)& \textbf{94.8}& 22135& \textbf{92.4}& 3582& \textbf{70.0}& 79367& 43.3& 84342& \textbf{95.0}& 40511\\
Self-Consistency ($N=8$)& 93.4& 22135& 90.1& 3582& 60.0& 79367& 40.0& 84342& 85.0& 40511\\
MR-Thinking ($N=8$)& 91.2& 21396& 88.4& 4792& 56.7& 78432& 40.0& 86568& 87.5& 36780\\
\textsc{OptScale}$^0$ ($N=8$)& \textbf{94.8}& \textbf{11354}& \textbf{92.4}& \textbf{1687}& \textbf{70.0}& \textbf{49505}& 43.3& 78803& \textbf{95.0}& \textbf{29288}\\
\textsc{OptScale}$^t$ ($N=8$)& \textbf{94.8}& 18236& \textbf{92.4}& 3492& \textbf{70.0}& 53855& \textbf{46.7}& \textbf{69661}& \textbf{95.0}& 30671\\
\midrule
Best-of-N (BoN) ($N=16$)& 93.8& 43909& \textbf{93.5}& 7160& \textbf{76.7}& 159479& 46.7& 168080& \textbf{95.0}& 79881\\
Self-Consistency ($N=16$)& 93.4& 43909& 90.8& 7160& 66.7& 159479& 36.7& 168080& 87.5& 79881\\
MR-Thinking ($N=16$)& 91.6& 42422& 88.3& 9615& 53.3& 157507& 43.3& 171853& 87.5& 72877\\
\textsc{OptScale}$^0$ ($N=16$)& \textbf{95.2}& \textbf{18123}& \textbf{93.5}& \textbf{4372}& \textbf{76.7}& \textbf{122001}& \textbf{50.0}& \textbf{126874}& \textbf{95.0}& 52310\\
\textsc{OptScale}$^t$ ($N=16$)& 94.4& 22487& \textbf{93.5}& 6860& \textbf{76.7}& 122896& \textbf{50.0}& 143181& \textbf{95.0}& \textbf{47061}\\
\midrule
Best-of-N (BoN) ($N=32$)& 94.4& 87160& 93.1& 14270& 70.0& 317453& \textbf{50.0}& 338555& \textbf{95.0}& 157962\\
Self-Consistency ($N=32$)& 93.2& 87160& 90.9& 14270& 63.3& 317453& 40.0& 338555& 90.0& 157962\\
MR-Thinking ($N=32$)& 91.4& 84234& 88.4& 19135& 63.3& 313463& 33.3& 341517& 90.0& 143178\\
\textsc{OptScale}$^0$ ($N=32$)& \textbf{95.0}& \textbf{34732}& \textbf{93.5}& \textbf{4757}& 73.3& 256177& \textbf{50.0}& \textbf{291402}& \textbf{95.0}& 66086\\
\textsc{OptScale}$^t$ ($N=32$)& \textbf{95.0}& 35317& \textbf{93.5}& 6337& \textbf{76.7}& \textbf{221290}& \textbf{50.0}& 304676& \textbf{95.0}& \textbf{65792}\\
\midrule
Best-of-N (BoN) ($N=64$)& 94.0& 174693& 94.0& 28547& 80.0& 637293& \textbf{53.3}& 676533& \textbf{95.0}& 312241\\
Self-Consistency ($N=64$)& 93.4& 174693& 91.2& 28547& 60.0& 637293& 40.0& 676533& 92.5& 312241\\
MR-Thinking ($N=64$)& 92.0& 168331& 88.2& 38305& 70.0& 625850& 40.0& 684446& 90.0& 286639\\
\textsc{OptScale}$^0$ ($N=64$)& \textbf{94.6}& 110001& \textbf{94.1}& \textbf{11086}& \textbf{83.3}& 503002& 50.0& \textbf{549344}& \textbf{95.0}& 119777\\
\textsc{OptScale}$^t$ ($N=64$)& \textbf{94.6}& \textbf{76284}& 94.0& 21386& 80.0& \textbf{343491}& \textbf{53.3}& 649900& \textbf{95.0}& \textbf{119282}\\

\midrule
\midrule
\multicolumn{11}{l}{\textbf{Llama-3.1-8B-Instruct}} \\
\midrule
Best-of-N (BoN) ($N=8$)& 63.6& 9609& \textbf{88.4}& 2160& \textbf{10.0}& 54212& \textbf{3.3}& 51510& \textbf{32.5}& 15321\\
Self-Consistency ($N=8$)& 58.6& 9609& 87.5& 2160& 3.3& 54212& 0.0& 51510& 27.5& 15321\\
MR-Thinking ($N=8$)& 41.6& 14918& 61.6& 2812& 0.0& 46037& \textbf{3.3}& \textbf{39374}& 12.5& 24814\\
\textsc{OptScale}$^0$ ($N=8$)& \textbf{63.8}& \textbf{8756}& \textbf{88.4}& \textbf{1462}& \textbf{10.0}& \textbf{42859}& \textbf{3.3}& 46970& \textbf{32.5}& 15113\\
\textsc{OptScale}$^t$ ($N=8$)& 63.6& 9479& \textbf{88.4}& 2136& \textbf{10.0}& 43212& \textbf{3.3}& 46724& \textbf{32.5}& \textbf{14323}\\
\midrule
Best-of-N (BoN) ($N=16$)& 66.2& 19067& 89.1& 4182& \textbf{10.0}& 108067& \textbf{0.0}& 105305& 37.5& 31319\\
Self-Consistency ($N=16$)& 59.6& 19067& 88.6& 4182& 6.7& 108067& \textbf{0.0}& 105305& 30.0& 31319\\
MR-Thinking ($N=16$)& 43.2& 30507& 62.1& 5723& 6.7& 92594& \textbf{0.0}& 85057& 15.0& 49970\\
\textsc{OptScale}$^0$ ($N=16$)& \textbf{67.4}& \textbf{16671}& 89.1& \textbf{2783}& \textbf{10.0}& 66126& \textbf{0.0}& 78369& 37.5& \textbf{28790}\\
\textsc{OptScale}$^t$ ($N=16$)& 66.8& 17311& \textbf{89.4}& 3613& \textbf{10.0}& \textbf{55707}& \textbf{0.0}& \textbf{72258}& \textbf{40.0}& 29498\\
\midrule
Best-of-N (BoN) ($N=32$)& 66.8& 37622& 89.4& 8419& \textbf{10.0}& 217408& \textbf{0.0}& 194240& 40.0& 61599\\
Self-Consistency ($N=32$)& 61.2& 37622& 88.6& 8419& 6.7& 217408& \textbf{0.0}& 194240& 32.5& 61599\\
MR-Thinking ($N=32$)& 41.4& 60396& 58.8& 11543& 3.3& 190111& \textbf{0.0}& 168523& 17.5& 93910\\
\textsc{OptScale}$^0$ ($N=32$)& 66.6& 35064& \textbf{89.5}& 5313& \textbf{10.0}& 125491& \textbf{0.0}& 90585& \textbf{42.5}& \textbf{44866}\\
\textsc{OptScale}$^t$ ($N=32$)& \textbf{67.2}& \textbf{30156}& \textbf{89.5}& \textbf{4201}& \textbf{10.0}& \textbf{110728}& \textbf{0.0}& \textbf{84817}& \textbf{42.5}& 46213\\
\midrule
Best-of-N (BoN) ($N=64$)& 68.8& 70643& 89.3& 14697& \textbf{13.3}& 386783& \textbf{0.0}& 348361& 42.5& 113401\\
Self-Consistency ($N=64$)& 60.8& 70643& 89.2& 14697& 6.7& 386783& \textbf{0.0}& 348361& 32.5& 113401\\
MR-Thinking ($N=64$)& 40.8& 122227& 60.7& 23423& 0.0& \textbf{377475}& \textbf{0.0}& 327074& 15.0& 189239\\
\textsc{OptScale}$^0$ ($N=64$)& \textbf{69.0}& 64179& 89.5& \textbf{5720}& \textbf{13.3}& 386772& \textbf{0.0}& 141904& \textbf{45.0}& \textbf{100675}\\
\textsc{OptScale}$^t$ ($N=64$)& 68.6& \textbf{60692}& \textbf{89.6}& 12044& \textbf{13.3}& 386668& \textbf{0.0}& \textbf{140707}& \textbf{45.0}& 106039\\
\bottomrule
\end{tabular}
}
\caption{Comparison of different inference-time scaling methods on common mathematical reasoning benchmarks (with $N=8$ and $N=64$). ``Acc.'' denotes accuracy (\%), ``Toks.'' indicate the total number of inference tokens.}
\label{tab:prepare_Distill-Qwen-7B}
\end{table*}

\begin{table*}[t!]
\centering
\resizebox{0.98\textwidth}{!}{%
\begin{tabular}{l c c c c c c c c c c}
\toprule
\multirow{2}{*}{\textbf{Baseline Method}} & \multicolumn{2}{c}{\textbf{MATH-500}} & \multicolumn{2}{c}{\textbf{GSM8K}} & \multicolumn{2}{c}{\textbf{AIME 2024}} & \multicolumn{2}{c}{\textbf{AIME 2025}} & \multicolumn{2}{c}{\textbf{AMC 2023}} \\ 
\cmidrule{2-11}
 & \textbf{Acc.}  & \textbf{Toks.} ($\downarrow$) & \textbf{Acc.} & \textbf{Toks.} ($\downarrow$) & \textbf{Acc.}  & \textbf{Toks.} ($\downarrow$)  & \textbf{Acc.} & \textbf{Toks.} ($\downarrow$) & \textbf{Acc.}  & \textbf{Toks.} ($\downarrow$) \\
\midrule
\multicolumn{11}{l}{\textbf{Qwen/QwQ-32B}} \\ 
\midrule
Best-of-N (BoN) ($N=8$)& 94.6& 30601& 95.8& 11177& \textbf{73.3}& 85853& \textbf{66.7}& 95235& \textbf{95.0}& 55291\\
Self-Consistency ($N=8$)& \textbf{95.2}& 30601& 95.6& 11177& 70.0& 85853& \textbf{66.7}& 95235& 92.5& 55291\\
MR-Thinking ($N=8$)& 94.6& 30378& 95.3& 12832& 70.0& 85864& 53.3& 95536& 87.5& 47288\\
\textsc{OptScale}$^0$ ($N=8$)& 95.0& \textbf{17469}& \textbf{95.9}& \textbf{5031}& \textbf{73.3}& \textbf{61449}& \textbf{66.7}& 85669& \textbf{95.0}& 42177\\
\textsc{OptScale}$^t$ ($N=8$)& 94.8& 26910& 95.8& 10767& \textbf{73.3}& 64094& \textbf{66.7}& \textbf{84766}& \textbf{95.0}& \textbf{36494}\\
\midrule
Best-of-N (BoN) ($N=16$)& 94.6& 61386& 95.5& 22372& \textbf{73.3}& 173326& 63.3& 191569& \textbf{97.5}& 110053\\
Self-Consistency ($N=16$)& \textbf{95.2}& 61386& 95.5& 22372& 66.7& 173326& 60.0& 191569& 92.5& 110053\\
MR-Thinking ($N=16$)& 94.4& 60801& 95.6& 26043& 70.0& 171256& 56.7& 189500& 80.0& 95434\\
\textsc{OptScale}$^0$ ($N=16$)& 94.6& \textbf{30425}& \textbf{95.8}& \textbf{8041}& \textbf{73.3}& 112828& \textbf{66.7}& 156993& \textbf{97.5}& \textbf{58604}\\
\textsc{OptScale}$^t$ ($N=16$)& 94.6& 38545& \textbf{95.8}& 14481& \textbf{73.3}& \textbf{112467}& \textbf{66.7}& \textbf{152568}& \textbf{97.5}& 59546\\
\midrule
Best-of-N (BoN) ($N=32$)& 95.2& 122847& 95.5& 44724& \textbf{73.3}& 347090& 63.3& 385336& \textbf{97.5}& 222667\\
Self-Consistency ($N=32$)& 95.0& 122847& 95.7& 44724& 70.0& 347090& 63.3& 385336& 92.5& 222667\\
MR-Thinking ($N=32$)& 94.4& 121565& 95.4& 52509& 63.3& 345355& 53.3& 383810& 75.0& 191141\\
\textsc{OptScale}$^0$ ($N=32$)& 95.2& \textbf{58051}& \textbf{95.8}& \textbf{15052}& \textbf{73.3}& 213113& 63.3& 306402& \textbf{97.5}& \textbf{111212}\\
\textsc{OptScale}$^t$ ($N=32$)& \textbf{95.4}& 72458& \textbf{95.8}& 29263& \textbf{73.3}& \textbf{210575}& \textbf{66.7}& \textbf{303369}& \textbf{97.5}& 117655\\
\midrule
Best-of-N (BoN) ($N=64$)& 94.8& 230402& 95.8& 83902& \textbf{76.7}& 649256& 66.7& 721405& 97.5& 420481\\
Self-Consistency ($N=64$)& 95.4& 230402& 95.8& 83902& 70.0& 649256& 63.3& 721405& 92.5& 420481\\
MR-Thinking ($N=64$)& 93.8& 227838& 94.8& 98607& 73.3& 646200& 56.7& 719654& 80.0& 360068\\
\textsc{OptScale}$^0$ ($N=64$)& \textbf{95.8}& 107720& \textbf{95.9}& \textbf{35985}& \textbf{76.7}& 399929& \textbf{70.0}& 556340& \textbf{100.0}& \textbf{190633}\\
\textsc{OptScale}$^t$ ($N=64$)& \textbf{95.8}& \textbf{106412}& 95.8& 42735& \textbf{76.7}& \textbf{375208}& \textbf{70.0}& \textbf{516346}& \textbf{100.0}& 202603\\

\midrule
\midrule
\multicolumn{11}{l}{\textbf{Deepseek-R1-Distill-Llama-8B}} \\
\midrule
Best-of-N (BoN) ($N=8$)& 82.0& 26497& 74.4& 8931& 43.3& 82784& 26.7& 59548& 75.0& 35598\\
Self-Consistency ($N=8$)& \textbf{88.4}& 26497& 81.8& 8931& \textbf{53.3}& 82784& 23.3& 59548& 77.5& 35598\\
MR-Thinking ($N=8$)& 85.4& \textbf{21868}& \textbf{90.5}& 8552& 43.3& 89923& \textbf{33.3}& 93307& \textbf{80.0}& 34500\\
\textsc{OptScale}$^0$ ($N=8$)& 81.8& 26363& 75.3& \textbf{7060}& 46.7& \textbf{64380}& 30.0& 58823& 75.0& 28212\\
\textsc{OptScale}$^t$ ($N=8$)& 82.0& 26119& 74.5& 8669& 43.3& 68296& 30.0& \textbf{58794}& \textbf{80.0}& \textbf{19026}\\
\midrule
Best-of-N (BoN) ($N=16$)& 83.0& 52784& 75.7& 17799& \textbf{50.0}& 161285& 30.0& 120587& 75.0& 72769\\
Self-Consistency ($N=16$)& \textbf{90.2}& 52784& 84.6& 17799& \textbf{50.0}& 161285& 23.3& 120587& 80.0& 72769\\
MR-Thinking ($N=16$)& 88.2& \textbf{43338}& \textbf{91.2}& 16916& \textbf{50.0}& 178390& \textbf{40.0}& 183332& \textbf{82.5}& 67480\\
\textsc{OptScale}$^0$ ($N=16$)& 83.4& 48246& 75.7& 17444& \textbf{50.0}& 125109& 33.3& \textbf{112037}& 80.0& \textbf{23955}\\
\textsc{OptScale}$^t$ ($N=16$)& 83.8& 50914& 76.8& \textbf{15746}& \textbf{50.0}& \textbf{124662}& 33.3& 113543& 77.5& 25738\\
\midrule
Best-of-N (BoN) ($N=32$)& 83.8& 105343& 76.3& 35601& 53.3& 324198& 30.0& 237661& 80.0& 143577\\
Self-Consistency ($N=32$)& \textbf{91.2}& 105343& 86.7& 35601& 50.0& 324198& 23.3& 237661& 75.0& 143577\\
MR-Thinking ($N=32$)& 86.4& \textbf{86127}& \textbf{90.9}& 33819& \textbf{60.0}& 356745& 26.7& 368740& \textbf{85.0}& 133628\\
\textsc{OptScale}$^0$ ($N=32$)& 84.2& 96210& 77.4& \textbf{32584}& 53.3& \textbf{245519}& \textbf{33.3}& \textbf{210819}& \textbf{85.0}& 104542\\
\textsc{OptScale}$^t$ ($N=32$)& 83.8& 104676& 76.4& 35014& 56.7& 267862& \textbf{33.3}& 211114& \textbf{85.0}& \textbf{98770}\\
\midrule
Best-of-N (BoN) ($N=64$)& 84.6& 210630& 76.9& 71237& \textbf{66.7}& 652661& 33.3& 476857& \textbf{82.5}& 287408\\
Self-Consistency ($N=64$)& \textbf{91.8}& 210630& 87.5& 71237& 56.7& 652661& 23.3& 476857& 80.0& 287408\\
MR-Thinking ($N=64$)& 88.2& \textbf{171565}& \textbf{90.8}& 67478& 56.7& 704771& 30.0& 742767& \textbf{82.5}& 266692\\
\textsc{OptScale}$^0$ ($N=64$)& 84.8& 176894& 78.2& \textbf{52563}& \textbf{66.7}& \textbf{400680}& \textbf{36.7}& \textbf{283022}& \textbf{82.5}& 197639\\
\textsc{OptScale}$^t$ ($N=64$)& 84.4& 192857& 77.6& 58729& \textbf{66.7}& 497003& \textbf{36.7}& 337419& 80.0& \textbf{168696}\\
\bottomrule
\end{tabular}
}
\caption{Comparison of different inference-time scaling methods on common mathematical reasoning benchmarks (with $N=8$ and $N=64$). ``Acc.'' denotes accuracy (\%), ``Toks.'' indicate the total number of inference tokens.}
\label{tab:prepare_QwQ-32B}
\end{table*}

\begin{table*}[t!]
\centering
\resizebox{0.98\textwidth}{!}{%
\begin{tabular}{l c c c c c c c c c c}
\toprule
\multirow{2}{*}{\textbf{Baseline Method}} & \multicolumn{2}{c}{\textbf{MATH-500}} & \multicolumn{2}{c}{\textbf{GSM8K}} & \multicolumn{2}{c}{\textbf{AIME 2024}} & \multicolumn{2}{c}{\textbf{AIME 2025}} & \multicolumn{2}{c}{\textbf{AMC 2023}} \\ 
\cmidrule{2-11}
 & \textbf{Acc.}  & \textbf{Toks.} ($\downarrow$) & \textbf{Acc.} & \textbf{Toks.} ($\downarrow$) & \textbf{Acc.}  & \textbf{Toks.} ($\downarrow$)  & \textbf{Acc.} & \textbf{Toks.} ($\downarrow$) & \textbf{Acc.}  & \textbf{Toks.} ($\downarrow$) \\
\midrule
\multicolumn{11}{l}{\textbf{Qwen3-8B}} \\ 
\midrule
Best-of-N (BoN) ($N=8$)& \textbf{91.0}& 32553& 95.6& 13655& \textbf{83.3}& 107101& \textbf{70.0}& 125791& \textbf{90.0}& 54416\\
Self-Consistency ($N=8$)& 89.4& 32553& 95.5& 13655& \textbf{83.3}& 107101& 66.7& 125791& 87.5& 54416\\
MR-Thinking ($N=8$)& 90.8& 31136& 95.5& 12464& 80.0& 101214& 66.7& 120220& \textbf{90.0}& 53909\\
\textsc{OptScale}$^0$ ($N=8$)& \textbf{91.0}& \textbf{19793}& \textbf{95.7}& \textbf{6128}& \textbf{83.3}& \textbf{89661}& \textbf{70.0}& \textbf{116109}& \textbf{90.0}& \textbf{36618}\\
\textsc{OptScale}$^t$ ($N=8$)& \textbf{91.0}& 28613& 95.6& 13268& \textbf{83.3}& 93424& \textbf{70.0}& 116413& \textbf{90.0}& 44023\\
\midrule
Best-of-N (BoN) ($N=16$)& 90.8& 65062& 95.8& 27268& \textbf{83.3}& 215354& \textbf{70.0}& 251192& \textbf{90.0}& 109692\\
Self-Consistency ($N=16$)& 89.6& 65062& 95.5& 27268& \textbf{83.3}& 215354& \textbf{70.0}& 251192& 85.0& 109692\\
MR-Thinking ($N=16$)& \textbf{92.0}& 61878& 95.5& 24811& 66.7& 203154& 66.7& 241374& \textbf{90.0}& 107545\\
\textsc{OptScale}$^0$ ($N=16$)& 90.8& \textbf{44287}& \textbf{95.9}& \textbf{10192}& \textbf{83.3}& 172566& \textbf{70.0}& 226058& \textbf{90.0}& \textbf{69936}\\
\textsc{OptScale}$^t$ ($N=16$)& 90.8& 47301& 95.8& 18554& \textbf{83.3}& \textbf{167600}& \textbf{70.0}& \textbf{222997}& \textbf{90.0}& 74377\\
\midrule
Best-of-N (BoN) ($N=32$)& 91.8& 130230& \textbf{95.8}& 54480& \textbf{80.0}& 433309& 66.7& 500503& \textbf{90.0}& 220643\\
Self-Consistency ($N=32$)& 89.6& 130230& 95.6& 54480& \textbf{80.0}& 433309& \textbf{70.0}& 500503& 85.0& 220643\\
MR-Thinking ($N=32$)& 89.4& 123215& 95.1& 49409& 73.3& 411748& 63.3& 487852& 87.5& 213189\\
\textsc{OptScale}$^0$ ($N=32$)& 91.8& \textbf{63956}& \textbf{95.8}& \textbf{20523}& \textbf{80.0}& 327228& 66.7& 446103& \textbf{90.0}& \textbf{125427}\\
\textsc{OptScale}$^t$ ($N=32$)& \textbf{92.0}& 69712& \textbf{95.8}& 30057& \textbf{80.0}& \textbf{316460}& 66.7& \textbf{433291}& \textbf{90.0}& 126604\\
\midrule
Best-of-N (BoN) ($N=64$)& 91.4& 260304& \textbf{95.9}& 109154& 76.7& 862151& 60.0& 996858& 87.5& 438448\\
Self-Consistency ($N=64$)& 89.2& 260304& 95.7& 109154& \textbf{80.0}& 862151& \textbf{73.3}& 996858& 85.0& 438448\\
MR-Thinking ($N=64$)& 90.6& 246414& 95.5& 98416& 66.7& 812549& \textbf{73.3}& 974575& 87.5& 428166\\
\textsc{OptScale}$^0$ ($N=64$)& \textbf{92.2}& 121983& \textbf{95.9}& \textbf{54398}& 76.7& 606271& 63.3& 899392& \textbf{90.0}& \textbf{245864}\\
\textsc{OptScale}$^t$ ($N=64$)& 92.0& \textbf{121975}& \textbf{95.9}& 72669& 76.7& \textbf{550789}& 63.3& \textbf{878513}& \textbf{90.0}& 252928\\

\midrule
\midrule
\multicolumn{11}{l}{\textbf{Llama3.2-1B-Instruct}} \\
\midrule
Best-of-N (BoN) ($N=8$)& \textbf{41.0}& 3280& 76.6& 2143& \textbf{3.3}& 14916& \textbf{0.0}& 15849& 15.0& 8396\\
Self-Consistency ($N=8$)& 29.0& 3280& 62.4& 2143& 0.0& 14916& \textbf{0.0}& 15849& 15.0& 8396\\
MR-Thinking ($N=8$)& 26.0& 5475& 33.3& 2058& \textbf{3.3}& \textbf{9507}& \textbf{0.0}& \textbf{9866}& \textbf{20.0}& \textbf{7614}\\
\textsc{OptScale}$^0$ ($N=8$)& \textbf{41.0}& \textbf{3210}& \textbf{76.7}& \textbf{1685}& \textbf{3.3}& 12845& \textbf{0.0}& 14710& 17.5& 8099\\
\textsc{OptScale}$^t$ ($N=8$)& \textbf{41.0}& 3252& 76.6& 2135& \textbf{3.3}& 13334& \textbf{0.0}& 13458& 17.5& 8290\\
\midrule
Best-of-N (BoN) ($N=16$)& \textbf{44.0}& 6557& 80.4& 4286& \textbf{3.3}& 29904& \textbf{0.0}& 31824& 20.0& 19528\\
Self-Consistency ($N=16$)& 30.4& 6557& 64.4& 4286& \textbf{3.3}& 29904& \textbf{0.0}& 31824& 15.0& 19528\\
MR-Thinking ($N=16$)& 24.8& 10927& 33.6& 4175& 0.0& \textbf{18458}& \textbf{0.0}& \textbf{18085}& 12.5& \textbf{15129}\\
\textsc{OptScale}$^0$ ($N=16$)& \textbf{44.0}& \textbf{6051}& \textbf{80.5}& \textbf{3347}& \textbf{3.3}& 22019& \textbf{0.0}& 24879& \textbf{22.5}& 19143\\
\textsc{OptScale}$^t$ ($N=16$)& \textbf{44.0}& 6135& 80.4& 4068& \textbf{3.3}& 21329& \textbf{0.0}& 22204& \textbf{22.5}& 17759\\
\midrule
Best-of-N (BoN) ($N=32$)& 46.2& 13107& \textbf{83.4}& 8576& \textbf{3.3}& 58649& 0.0& 63609& 17.5& 42408\\
Self-Consistency ($N=32$)& 29.8& 13107& 65.4& 8576& \textbf{3.3}& 58649& 0.0& 63609& 10.0& 42408\\
MR-Thinking ($N=32$)& 24.2& 21869& 32.0& 8451& 0.0& 36468& \textbf{3.3}& 36831& 10.0& 31353\\
\textsc{OptScale}$^0$ ($N=32$)& \textbf{47.0}& \textbf{11444}& \textbf{83.4}& \textbf{6247}& \textbf{3.3}& 35344& 0.0& 40544& 17.5& 41926\\
\textsc{OptScale}$^t$ ($N=32$)& 46.8& 11571& \textbf{83.4}& 7231& \textbf{3.3}& \textbf{33964}& 0.0& \textbf{34406}& \textbf{20.0}& \textbf{28263}\\
\midrule
Best-of-N (BoN) ($N=64$)& 48.4& 26110& \textbf{86.4}& 17154& \textbf{3.3}& 118462& \textbf{0.0}& 123878& \textbf{20.0}& 86337\\
Self-Consistency ($N=64$)& 30.2& 26110& 66.7& 17154& \textbf{3.3}& 118462& \textbf{0.0}& 123878& 12.5& 86337\\
MR-Thinking ($N=64$)& 23.8& 43702& 32.2& 17005& 0.0& 74689& \textbf{0.0}& 73375& 7.5& \textbf{63126}\\
\textsc{OptScale}$^0$ ($N=64$)& \textbf{48.8}& 23838& \textbf{86.4}& \textbf{10700}& \textbf{3.3}& 50778& \textbf{0.0}& 45345& \textbf{20.0}& 82040\\
\textsc{OptScale}$^t$ ($N=64$)& \textbf{48.8}& \textbf{23827}& \textbf{86.4}& 12377& \textbf{3.3}& \textbf{49967}& \textbf{0.0}& \textbf{42545}& \textbf{20.0}& 82946\\
\bottomrule
\end{tabular}
}
\caption{Comparison of different inference-time scaling methods on common mathematical reasoning benchmarks (with $N=8$ and $N=64$). ``Acc.'' denotes accuracy (\%), ``Toks.'' indicate the total number of inference tokens.}
\label{tab:prepare_Qwen3-8B}
\end{table*}

\begin{table*}[t!]
\centering
\resizebox{0.7\textwidth}{!}{%
\setlength{\tabcolsep}{4pt}
\begin{tabular}{@{}cccccccccc@{}}
\toprule
\multirow{2}{*}{\textbf{Temperature}} & \multirow{2}{*}{\textbf{Method}} & \multicolumn{2}{c}{\textbf{\textit{N}=8}} & \multicolumn{2}{c}{\textbf{\textit{N}=16}} & \multicolumn{2}{c}{\textbf{\textit{N}=32}} & \multicolumn{2}{c}{\textbf{\textit{N}=64}} \\
\cmidrule(lr){3-4} \cmidrule(lr){5-6} \cmidrule(lr){7-8} \cmidrule(lr){9-10}
& & \textbf{Acc.} & \textbf{Toks.} ($\downarrow$) & \textbf{Acc.} & \textbf{Toks.} ($\downarrow$) & \textbf{Acc.} & \textbf{Toks.} ($\downarrow$) & \textbf{Acc.} & \textbf{Toks.} ($\downarrow$) \\
\midrule
\multirow{2}{*}{0.2} & BoN & 70.0 & 84K & 70.0 & 169K & 70.0 & 338K & 73.3 & 671K \\
                     & OptScale$^0$ & 70.0 & 56K & 76.7 & 150K & 73.3 & 157K & 73.3 & 288K \\
\midrule
\multirow{2}{*}{0.6} & BoN & 70.0 & 79K & 76.7 & 159K & 70.0 & 317K & 80.0 & 637K \\
                     & OptScale$^0$ & 70.0 & 50K & 76.7 & 122K & 73.3 & 255K & 83.3 & 504K \\
\midrule
\multirow{2}{*}{0.8} & BoN & 66.7 & 82K & 73.3 & 161K & 70.0 & 321K & 70.0 & 648K \\
                     & OptScale$^0$ & 66.7 & 60K & 73.3 & 128K & 73.3 & 186K & 73.3 & 281K \\
\midrule
\multirow{2}{*}{1.0} & BoN & 63.3 & 84K & 70.0 & 166K & 70.0 & 329K & 70.0 & 659K \\
                     & OptScale$^0$ & 70.0 & 67K & 76.7 & 120K & 73.3 & 208K & 70.0 & 538K \\
\bottomrule
\end{tabular}}
\caption{Comparison of different decoding temperatures (i.e., different levels of correlations) for running Deepseek-R1-Distill-Qwen-7B on AIME24. ``Acc.'' denotes accuracy (\%), ``Toks.'' indicate the total number of inference tokens in thousands.}
\label{tab:iid-table}
\end{table*}

\paragraph{Robustness to Distribution Approximation.}

While our method models verifier score distributions using a truncated normal approximation, the theoretical guarantees rely solely on empirical mean and variance. To assess the sensitivity of our approach to modeling inaccuracies, we selected 30 questions from various datasets where this approximation exhibited the poorest fit.
Despite the distributional mismatch, \textsc{OptScale} achieves a 2.5\% accuracy gain and a 48.8\% reduction in token usage on this challenging subset, demonstrating the algorithm's robustness even under significant deviations from the assumed distribution.

\paragraph{Robustness to Different Verifiers.} 

Our primary experiments use Qwen2.5-Math-PRM-7B as the verifier. To assess the generality of \textsc{OptScale} across different verifiers, we further used Math-Shepherd-Mistral-7B-PRM as the verifier. Table~\ref{tab:mistral-prm-table} reports the full results for Deepseek-R1-Distill-Qwen-7B across all benchmarks.
Across all five benchmarks, \textsc{OptScale} continues to deliver consistent gains, achieving an average of a 0.9\% accuracy improvement and 16.4\% token reduction.
This demonstrates that our \textsc{OptScale} is robust to different verifiers.

\begin{table*}[t!]
\centering
\resizebox{0.97\textwidth}{!}{%
\begin{tabular}{l c c c c c c c c c c}
\toprule
\multirow{2}{*}{\textbf{Baseline Method}} & \multicolumn{2}{c}{\textbf{MATH-500}} & \multicolumn{2}{c}{\textbf{GSM8K}} & \multicolumn{2}{c}{\textbf{AIME 2024}} & \multicolumn{2}{c}{\textbf{AIME 2025}} & \multicolumn{2}{c}{\textbf{AMC 2023}} \\ 
\cmidrule{2-11}
 & \textbf{Acc.}  & \textbf{Toks.} ($\downarrow$) & \textbf{Acc.} & \textbf{Toks.} ($\downarrow$) & \textbf{Acc.}  & \textbf{Toks.} ($\downarrow$)  & \textbf{Acc.} & \textbf{Toks.} ($\downarrow$) & \textbf{Acc.}  & \textbf{Toks.} ($\downarrow$) \\
\midrule
\multicolumn{11}{l}{\textbf{Deepseek-R1-Distill-Qwen-7B} (w/ Math-Shepherd-Mistral-7B-PRM as the verifier)} \\ 
\midrule
Best-of-N (BoN) ($N=8$)& 92.4 & 22135 & 89.7 & 3582 & 63.3 & 79367 & 50.0 & 84342 & 90.0 & 40511 \\
\textsc{OptScale}$^0$ (Ours) ($N=8$)& 92.4& 21326& 90.0& 2357& 63.3& 78287& 50.0& 84342& 90.0& 38968\\
\midrule
Best-of-N (BoN) ($N=16$)& 91.4 & 43909 & 90.8 & 7160 &63.3 & 159479 & 53.3 &  168080 & 90.0 & 79881 \\
\textsc{OptScale}$^0$ (Ours) ($N=16$)& 91.6& 40931& 90.8& 4187& 66.7& 155320& 53.3& 166073& 90.0&72692\\
\midrule
Best-of-N (BoN) ($N=32$)& 92.0& 87160& 91.0& 14270&66.7& 317453& 53.3&  338555& 90.0& 157962\\
\textsc{OptScale}$^0$ (Ours) ($N=32$)& 92.0& 76787& 91.1& 10225& 70.0& 287129& 53.3& 324314& 90.0&125625\\
\midrule
Best-of-N (BoN) ($N=64$)& 91.4& 174693& 91.3& 28547&63.3& 637293& 46.7&  676533& 92.5& 312241\\
\textsc{OptScale}$^0$ (Ours) ($N=64$)& 91.4& 165798& 91.5& 12532& 66.7& 457143& 50.0& 501845& 92.5&245073\\
\bottomrule
\end{tabular}
}
\caption{Performance on common reasoning benchmarks using Math-Shepherd-Mistral-7B-PRM as the verifier ($N\in \{8, 16, 32, 64\}$). ``Acc.'' denotes accuracy (\%), ``Toks.'' indicate the total number of inference tokens.}
\label{tab:mistral-prm-table}
\end{table*}

\paragraph{Generalization Beyond Mathematical Reasoning.}

To evaluate cross-domain generality, we applied \textsc{OptScale} to GPQA, a challenging question-answering benchmark. 
Results summarized in \ref{tab:appendix-gpqa-table} show that with Qwen and LLaMA backbones, \textsc{OptScale} achieves \textbf{+5.3\%} and \textbf{+4.6\%} accuracy gains, while reducing token usage by \textbf{13.6\%} and \textbf{2.5\%}, respectively.
These findings indicate that the probabilistic scaling principle underlying \textsc{OptScale} generalizes effectively beyond mathematical reasoning tasks.

\begin{table*}[t!]
\centering
\resizebox{0.64\textwidth}{!}{%
\setlength{\tabcolsep}{5.5pt}
\begin{tabular}{@{}lcccccccc@{}}
\toprule
\multirow{2}{*}{\textbf{Method}} & \multicolumn{2}{c}{\textbf{\textit{N}=8}} & \multicolumn{2}{c}{\textbf{\textit{N}=16}} & \multicolumn{2}{c}{\textbf{\textit{N}=32}} & \multicolumn{2}{c}{\textbf{\textit{N}=64}} \\
\cmidrule(lr){2-3} \cmidrule(lr){4-5} \cmidrule(lr){6-7} \cmidrule(lr){8-9}
& \textbf{Acc.} & \textbf{Toks.} ($\downarrow$) & \textbf{Acc.} & \textbf{Toks.} ($\downarrow$) & \textbf{Acc.} & \textbf{Toks.} ($\downarrow$) & \textbf{Acc.} & \textbf{Toks.} ($\downarrow$) \\
\midrule
\multicolumn{9}{l}{\textbf{Llama-3.1-8B-Instruct}} \\
\midrule
BoN      & 13.1 & 14.6K  & 12.1 & 28.9K & 13.1 & 57.2K & 14.1 & 113.7K \\
OptScale$^0$ & 14.1 & 13.6K  & 14.1 & 25.8K & 13.1 & 57.0K & 15.7 & 112.6K \\
\midrule
\multicolumn{9}{l}{\textbf{Deepseek-R1-Distill-Qwen-7B}} \\
\midrule
BoN      & 5.1 & 54.1K  & 3.5 & 108.0K & 3.5 & 216.8K & 4.0 & 432.8K \\
OptScale$^0$ & 6.1 & 45.7K  & 5.6 & 82.7K & 4.6 & 149.0K & 83.3 & 423.7K \\
\bottomrule
\end{tabular}}
\caption{Validation of \textsc{OptScale}'s performance on GPQA benchmark across different sampling budgets ($N$).  ``Acc.'' denotes accuracy (\%), ``Toks.'' indicate the total number of inference tokens in thousands.}
\label{tab:appendix-gpqa-table}
\end{table*}

\subsection{Case Studies}
\label{sec: case_studies}

We present some representative cases to demonstrate the advantage of \textsc{OptScale} in terms of both performance and token consumption compared to Best-of-N (BoN), one of the most representative parallel inference-time scaling methods.

\textbf{Case 1} (See Table~\ref{tab:case1}) is a representative of very easy questions. Both BoN and \textsc{OptScale} start with a correct answer in the first round. However, BoN introduces an overwhelmingly additional wasted rounds because of the fixed N required. And the model is distracted by the small number of completions with wrong answers it generates across. Overall, BoN wasted 42 completions as the final selected answer actually comes from Completion 18. In contrast, \textsc{OptScale} has successfully identified that this is a very easy question, and stops reasoning at the 10th Completion, resulting in significant token consumption reduction of 80.13\%.

\textbf{Case 2} (See Table~\ref{tab:case2}) is a representative of difficult questions. Both BoN and \textsc{OptScale} hit the correct answer for the first time on Completion 4. During the entire process, lots of different answers are generated, of course, among which only 1 answer is correct. \textsc{OptScale} successfully recognizes the proper level of difficulty for this question and stops reasoning at Completion 22 through the dynamic adjustment mechanism. However, BoN introduces additional rounds because of its nature of generating a fixed number of rounds for each question, the model seems to be confused by various wrong answers generated during the excess sampled generations. Eventually, such confusion results in BoN choosing the wrong answer in the end. \textsc{OptScale} achieves a significant reduction of token consumption of 65.14\% while better utilizing this scaled compute for achieving the final correctness for this question.

\textbf{Case 3} (See Table~\ref{tab:case3}) is a representative of questions with middle-level difficulty, typically on the secondary school Mathematics Competition level. Both BoN and \textsc{OptScale} hit the correct answer at Completion 1. Most of the time, the model could generate the correct answer with minor wrong answer distractions. However, BoN introduces an additional round, making it waste all the generations from Completion 14 to Completion 60 as no higher score is assigned (the final generation BoN chooses is at Completion 13). \textsc{OptScale} terminates on Completion 
19, significantly reducing token consumption by 69.47\% by correctly evaluating how much inference-time compute to allocate for this question. Both methods answer this question correctly in the end.

\textbf{Case 4} (See Table~\ref{tab:case4}) is another representative of a difficult question. Both BoN and \textsc{OptScale} hit the correct answer at Completion 3 for the first time. Similar to Case 2, some of the generated answers are wrong across the different sampled completions. BoN introduces additional completions, which creates confusion for itself as more wrong answers come in during the later phase of sampled generations. Such confusion eventually causes BoN to select the wrong answer. \textsc{OptScale} decides when to terminate on Completion 14 appropriately, reducing token consumption by 76.07\% and obtaining the correct final answer.

\textbf{Case 5} (See Table~\ref{tab:case5}) is a representative of questions of upper-middle level difficulty. Both BoN and \textsc{OptScale} hit the correct answer at Completion 2 for the first time. The majority of the answers are correct. BoN introduces additional completions to answer this question correctly, while \textsc{OptScale} successfully evaluates this question and decides to terminate on Completion 6. This proves to be clever, as the final chosen Completion is also Completion 6, achieving completely no waste on inference-time compute scaling. This helps \textsc{OptScale} significantly reduce token consumption by 90.04\% and obtain the correct final answer.

\onecolumn
\begin{longtable}{p{3.5cm}|p{9cm}|p{4cm}}
\caption{Case 1 (Dataset: GSM8K, Backbone Model: Llama-3.2-1B-Instruct).} 
\label{tab:case1} 
\\
\toprule
\multicolumn{3}{>{\columncolor[gray]{0.95}}p{0.96\textwidth}<{\arraybackslash}}{
  \textbf{Question:} \textcolor{blue}{A robe takes 2 bolts of blue fiber and half that much white fiber. How many bolts in total does it take?}
} \\
\midrule
\textbf{N} & \textbf{Best-of-N (BoN)} & \textsc{\textbf{OptScale}} \\
\midrule
\endfirsthead

\toprule
\multicolumn{3}{>{\columncolor[gray]{0.95}}p{0.96\textwidth}<{\arraybackslash}}{
  \textbf{(Continued) Question:} \textcolor{blue}{A robe takes 2 bolts of blue fiber and half that much white fiber. How many bolts in total does it take?}
} \\
\midrule
\textbf{N} & \textbf{Best-of-N (BoN)} & \textsc{\textbf{OptScale}} \\
\midrule
\endhead

\midrule
\multicolumn{3}{c}{\textit{Continued on next page}} \\
\bottomrule
\endfoot

\bottomrule
\endlastfoot

\textbf{Completion 1} & \textit{... \# Step 3: Calculate the total amount of fiber required.} \newline The final answer is: \textcolor{blue}{$\boxed{3}$}  \newline (Already hit the correct answer) & Same as BoN \\
\midrule
\textbf{Completion 2} & \textit{... total\_fiber\_needed = blue\_fiber\_needed + white\_fiber\_needed = 4 + 2 = 6 ...} \newline The final answer is: \textcolor{red}{$\boxed{6}$} \newline (Move to the wrong answer) & Same as BoN \\
\midrule
\textbf{Completion 3-9} & (The final answer is kept at the correct answer ``3''.) & Same as BoN \\
\midrule
\textbf{Completion 10} & (BoN keeps on generating new completions as it doesn't know when to stop.) & Reaches the optimal number of sampled completions. \textsc{OptScale} terminates here with the correct answer 3. \\
\midrule
\textbf{Completion 22} & (BoN jumps back to wrong answer ``6''.) & \\
\midrule
\textbf{Completion 43} & \textit{... \# Step 3: Calculate the total amount of fiber required. Total fiber required is 2 bolts of blue fiber + 1/2 * white fiber.} \newline \newline \textit{\# Step 4: Simplify the expression. To simplify the expression, we can convert 2 bolts of blue fiber to 2 * 2 = 4 bolts, since 2 bolts * 2 = 4. Then the expression becomes 4 + 1/2 * white fiber.} \newline \newline The final answer is: \textcolor{red}{$\boxed{4.25}$} \newline (BoN has a wrong understanding of this easy question and moves to a wrong answer ``4.25''.) & \\
\midrule
\textbf{Completion 60} & (BoN eventually stops as required when N=60 and gives the correct answer 3 after choosing the answer with the highest verifier score.) & \\
\midrule
\textbf{Token Consumption} & 12,103 &  2,485 \\
\midrule
\textbf{Correctness} &  \textcolor{blue}{Yes (\ding{51})}  &  \textcolor{blue}{Yes (\ding{51})} \\
\bottomrule
\end{longtable}

\onecolumn
\begin{longtable}{p{3.5cm}|p{10cm}|p{3.8cm}}
\caption{Case 2 (Dataset: AIME24, Backbone Model: Deepseek-R1-Distill-Qwen-7B).} 
\label{tab:case2} 
\\
\toprule
\multicolumn{3}{>{\columncolor[gray]{0.95}}p{0.96\textwidth}<{\arraybackslash}}{
\textbf{Question:} \textcolor{blue}{A list of positive integers has the following properties:
$\bullet$ The sum of the items in the list is $30$.
$\bullet$ The unique mode of the list is $9$.
$\bullet$ The median of the list is a positive integer that does not appear in the list itself.
Find the sum of the squares of all the items in the list.}
} \\
\midrule
\textbf{N} & \textbf{Best-of-N (BoN)} & \textsc{\textbf{OptScale}} \\
\midrule
\endfirsthead

\toprule
\multicolumn{3}{>{\columncolor[gray]{0.95}}p{0.96\textwidth}<{\arraybackslash}}{
\textbf{(Continued) Question:} \textcolor{blue}{A list of positive integers has the following properties:
$\bullet$ The sum of the items in the list is $30$.
$\bullet$ The unique mode of the list is $9$.
$\bullet$ The median of the list is a positive integer that does not appear in the list itself.
Find the sum of the squares of all the items in the list.}
} \\
\midrule
\textbf{N} & \textbf{Best-of-N (BoN)} & \textsc{\textbf{OptScale}} \\
\midrule
\endhead

\midrule
\multicolumn{3}{c}{\textit{Continued on next page}} \\
\bottomrule
\endfoot

\bottomrule
\endlastfoot

\textbf{Completion 4} & \textit{... After several attempts and checking for valid configurations, we find that the list \([5, 7, 9, 9]\) satisfies all conditions:}
\newline
\textit{- The sum is \(5 + 7 + 9 + 9 = 30\).}
\newline
\textit{- The mode is 9, appearing twice, more than any other number.}
\newline
\textit{- The median is \(\frac{7 + 9}{2} = 8\), which is not in the list.}
\newline
\textit{Finally, we calculate the sum of the squares of all the items:}
\newline

\textit{$5^2$ + $7^2$ + $9^2$ + $9^2$ = 25 + 49 + 81 + 81 = 236}
\newline
\textit{Thus, the sum of the squares of all the items in the list is \(\textcolor{blue}{\boxed{236}}\).} 
\newline (The first time the models hit the correct answer) & Same as BoN \\
\midrule

\textbf{Completion 6} & \textit{... Constructing the list:}
\newline
\textit{- The mode is 9, which must appear at least twice.}
\newline
\textit{- The median is the average of the second and third numbers, which is an integer not present in the list.}
\newline
\textit{We find that the list [5, 7, 9, 9] satisfies all conditions: ...}
\newline
\textit{- Calculating the sum of the squares of the items in the list:}
\newline
\textit{- \(5^2 + 7^2 + 9^2 + 9^2 = 25 + 49 + 81 + 81 = 236\).}
\newline
\textit{Thus, the sum of the squares of all the items in the list is \(\textcolor{blue}{\boxed{236}}\).} \newline (Hit the correct answer again) & Same as BoN \\
\midrule

\textbf{Completion 8} & \textit{... After several attempts and considerations, we find that the list can be structured as follows:}
\newline
\textit{- The list has 6 elements, as this is the smallest odd number that allows the median to be a positive integer not in the list.}
\newline
\textit{- The list is sorted and includes two 9s, which are the mode.}
\newline
\textit{- The remaining four numbers must sum to 12 and be distinct positive integers different from 9.}
\newline
\textit{The valid list found is:}
\textit{\[ [1, 2, 4, 5, 9, 9] \]}
\textit{This list satisfies all conditions:}
\newline
\textit{- The sum is \( 1 + 2 + 4 + 5 + 9 + 9 = 30 \).}
\newline
\textit{- The mode is 9, appearing twice.}
\newline
\textit{- The median is the average of the 3rd and 4th numbers, which is \( \frac{5 + 9}{2} = 7 \), a positive integer not in the list.}
\newline
\textit{Finally, we calculate the sum of the squares of all the items:}
\newline
\textit{ $1^2$ + $2^2$ + $4^2$ + $5^2$ + $9^2$ + $9^2$ = 1 + 4 + 16 + 25 + 81 + 81 = 208 }
\newline
\textit{Thus, the sum of the squares of all the items in the list is:} \textcolor{red}{\boxed{208}}
\newline (Previously, all wrong answers couldn't be extracted as numbers. This is the first numerical wrong answer.) & Same as BoN \\
\midrule

\textbf{Completion 22} & \textit{... Wait, n=7: two 9s, sum contributed by 9s=18. Sum remaining=12.}
\newline
\textit{So, five numbers, each positive integer, sum=12, each number appears at most once.}
\newline
\textit{Minimal sum for five distinct positive integers:1+2+3+4+5=15>12. So, impossible.}
\newline
\textit{n=6: two 9s, sum=18. Sum remaining=12.}
\newline
\textit{n=6: four numbers, sum=12, each number appears at most once.}
\newline
\textit{So, minimal sum for four distinct positive integers:1+2+3+4} 
\newline
(BoN continues generation.) & (\textsc{OptScale} terminates. Based on verifier scoring, it chooses the answer from Iteration 6 as the final answer, which is correct.) \\
\midrule

\textbf{Completion 26-35} & \textit{We need to find the sum of the squares of all the items in the list.}
\newline
\textit{First, we consider the number of elements in the list, denoted as \( n \). Since 9 is the unique mode, it must appear at least twice. We also need to ensure the median is an integer not present in the list.}
\newline
\textit{After analyzing different possible values for \( n \), we find that \( n = 6 \) works. The list must have three 9s and three 1s to satisfy all conditions:}
\newline
\textit{- The sum of the list is \( 1 + 1 + 1 + 9 + 9 + 9 = 30 \).}
\newline
\textit{- The mode is 9, which appears three times, more than any other number.}
\newline
\textit{- The median is the average of the third and fourth numbers, which is \( \frac{1 + 9}{2} = 5 \), an integer not present in the list.}
\newline
\textit{The sum of the squares of the items in the list is calculated as follows:}
\newline
\textit{ $1^2$ + $1^2$ + $1^2$ + $9^2$ + $9^2$ + $9^2$ = 1 + 1 + 1 + 81 + 81 + 81 = 246 }
\newline
\textit{Thus, the sum of the squares of all the items in the list is \(\textcolor{red}{\boxed{246}}\).} 
\newline (BoN keeps struggling between right and wrong answers. Correct answers appear several times, while wrong answers, such as 201 and 246, also appear.) & \\
\midrule

\textbf{Completion 37} & \textit{\textbf{Final Answer}}
\newline
\textit{The sum of the squares of all the items in the list is \boxed{208}.}
\newline
\textit{</think>}
\newline
\textit{Given the properties of the list of positive integers:}
\newline
\textit{1. The sum of the items is 30.}
\newline
\textit{2. The unique mode is 9.}
\newline
\textit{3. The median is a positive integer that does not appear in the list.}
\newline
\textit{We need to find the sum of the squares of all the items in the list.}
\newline
\textit{First, we note that 9 is the unique mode, so it must appear at least twice. The sum of the list is 30, so the remaining sum after accounting for two 9s is 12.}
\newline
\textit{Considering the median, since the median must be a positive integer not in the list, the list must have an even number of elements. We explore the case with 6 elements (n=6), as it is the smallest even number that allows the median to be a non-integer.}
\newline
\textit{For n=6, the list has two 9s and four other distinct numbers summing to 12. The possible sets of four distinct numbers are [1, 2, 3, 6] and [1, 2, 4, 5]. The set [1, 2, 3, 6] results in a median that is an integer, which is invalid. The set [1, 2, 4, 5] results in a median that is not in the list.}
\newline
\textit{Thus, the list is [1, 2, 4, 5, 9, 9]. The sum of the squares of these numbers is calculated as follows:}
\newline
\textit{[}
\textit{$1^2$ + $2^2$ + $4^2$ + $5^2$ + $9^2$ + $9^2$ = 1 + 4 + 16 + 25 + 81 + 81 = 208}
\textit{]}
\newline
\textit{Therefore, the sum of the squares of all the items in the list is \textcolor{red}{\boxed{208}}.}
\newline (BoN generates the wrong answer 208 again.) & \\
\midrule

\textbf{Completion 38-59} & \textit{... After considering all possibilities, the only valid list is for \( n = 4 \) with the sum of squares calculated as follows:}
\newline
\textit{- The list is [5, 7, 9, 9].}
\newline
\textit{- The sum of the squares is \( 5^2 + 7^2 + 9^2 + 9^2 = 25 + 49 + 81 + 81 = 236 \).}
\newline
\textit{Thus, the sum of the squares of all the items in the list is \textcolor{blue}{\boxed{236}}.} \newline (During this period, BoN either generates the correct answer "236" or "None".) & \\
\midrule

\textbf{Completion 60} & \textit{After testing various configurations, we find that the list [5, 7, 9, 9] satisfies all conditions:}
\newline
\textit{- The sum is 5 + 7 + 9 + 9 = 30.}
\newline
\textit{- The mode is 9, which appears twice.}
\newline
\textit{- The median is (7 + 9)/2 = 8, which is not in the list.}
\newline
\textit{The sum of the squares of the items in the list is calculated as follows:}
\newline
\textit{$5^2$ + $7^2$ + $9^2$ + $9^2$ = 25 + 49 + 81 + 81 = 256}
\newline
\textit{Thus, the sum of the squares of all the items in the list is \boxed{256}.} \newline (BoN eventually stops as required when N=60. In Completion 60, the model generates the wrong answer "256". Unfortunately, BoN chooses the answer in Completion 37 based on PRM scoring, which is the wrong answer 208. Despite spending so many tokens, BoN fails to give the correct answer.) & \\
\midrule

\textbf{Token Consumption} & 544,797 & 189,908 \\
\midrule
\textbf{Correctness} & \textcolor{red}{No (\ding{55})} & \textcolor{blue}{Yes (\ding{51})} \\
\bottomrule
\end{longtable}

\onecolumn
\begin{longtable}{p{3.5cm}|p{10cm}|p{3.5cm}}
\caption{Case 3 (Dataset: AMC23, Backbone Model: Deepseek-R1-Distill-Qwen-7B).}
\label{tab:case3} \\
\toprule
\multicolumn{3}{>{\columncolor[gray]{0.95}}p{0.96\textwidth}<{\arraybackslash}}{
\textbf{Question:} \textcolor{blue}{For how many ordered pairs $(a,b)$ of integers does the polynomial $x^3+ax^2+bx+6$ have $3$ distinct integer roots?}
} \\
\midrule
\textbf{N} & \textbf{Best-of-N (BoN)} & \textsc{\textbf{OptScale}} \\
\midrule
\endfirsthead

\toprule
\multicolumn{3}{>{\columncolor[gray]{0.95}}p{0.96\textwidth}<{\arraybackslash}}{
\textbf{(Continued) Question:} \textcolor{blue}{For how many ordered pairs $(a,b)$ of integers does the polynomial $x^3+ax^2+bx+6$ have $3$ distinct integer roots?}
} \\
\midrule
\textbf{N} & \textbf{Best-of-N (BoN)} & \textsc{\textbf{OptScale}} \\
\midrule
\endhead

\midrule
\multicolumn{3}{c}{\textit{Continued on next page}} \\
\bottomrule
\endfoot

\bottomrule
\endlastfoot

\textbf{Completion 1} &
\begin{minipage}[t]{\linewidth}
\textit{To determine the number of ordered pairs \((a, b)\) of integers such that the polynomial \(x^3 + ax^2 + bx + 6\) has three distinct integer roots, we use the Rational Root Theorem and factorization.\\
The polynomial must have integer roots, which are divisors of the constant term 6. The possible roots are \(\pm 1, \pm 2, \pm 3, \pm 6\).\\
The polynomial can be expressed as \((x - r)(x - s)(x - t)\) where \(r, s, t\) are distinct integers. Expanding this gives:
\[
x^3 - (r + s + t)x^2 + (rs + rt + st)x - rst
\]
Comparing coefficients:
\[
a = -(r + s + t), \quad b = rs + rt + st, \quad rst = -6
\]
Each valid triple leads to a unique ordered pair \((a, b)\). The valid sets are:}
\[
\begin{aligned}
(-1, 2, 3) &\rightarrow (-4, 1) \\
(-2, 1, 3) &\rightarrow (-2, -5) \\
(-3, 1, 2) &\rightarrow (0, -7) \\
(-1, -2, -3) &\rightarrow (6, 11) \\
(6, 1, -1) &\rightarrow (-6, -1)
\end{aligned}
\]
\textit{Thus, the number of ordered pairs \((a,b)\) is \( \boxed{5} \).}
\end{minipage} &
Same as BoN \\
\midrule

\textbf{Completion 2-8} &
\textit{The correct answer is repeated in Iteration 3. Some answers in other iterations were not completed before reaching the max token limit.} &
Same as BoN \\
\midrule

\textbf{Completion 9} &
\begin{minipage}[t]{\linewidth}
\textit{We find all sets of three distinct integers whose product is \(-6\).\\
Possible sets: (-1,2,3), (-2,1,3), (-3,1,2), (-1,-2,-3).\\
Each gives a unique \((a,b)\), except only 4 are counted. Therefore the answer is \(\boxed{4}\). (Incorrect)}
\end{minipage} &
Same as BoN \\
\midrule

\textbf{Completion 10-18} &
\begin{minipage}[t]{\linewidth}
\textit{Most completions correctly derive the answer \( \boxed{5} \), though Completion 10 and Completion 16 repeat wrong answer \( \boxed{4} \).}
\end{minipage} &
Same as BoN \\
\midrule

\textbf{Completion 19} &
\begin{minipage}[t]{\linewidth}
\textit{Correct pairs: (-6, -1), (-4, 1), (-2, -5), (0, -7), (6, 11).\\
Hence, total is \(\boxed{5}\).}
\end{minipage} &
\textsc{OptScale} selects Completion 13  with the correct answer. \\
\midrule

\textbf{Completion 19-59} &
\textit{Most completions repeat correct answer \(\boxed{5}\). Completion 33 repeats wrong answer \(\boxed{4}\).} &
\\
\midrule

\textbf{Completion 60} &
\begin{minipage}[t]{\linewidth}
\textit{Final correct answer \(\boxed{5}\) is confirmed.\\
Best PRM scoring still from Completion 13.}
\end{minipage} &
\\
\midrule

\textbf{Token Consumption} & 534,991 & 163,311 \\
\midrule
\textbf{Correctness} & \textcolor{blue}{Yes (\ding{51})} & \textcolor{blue}{Yes (\ding{51})}  \\
\end{longtable}

\onecolumn
\begin{longtable}{p{3.5cm}|p{10cm}|p{3.5cm}}
\caption{Case 4 (Dataset: AIME25, Backbone Model: Deepseek-R1-Distill-Qwen-7B).} \label{tab:case4} \\
\toprule
\multicolumn{3}{>{\columncolor[gray]{0.95}}p{0.96\textwidth}<{\arraybackslash}}{
\textbf{Question:} \textcolor{blue}{Sixteen chairs are arranged in a row. Eight people each select a chair in which to sit so that no person sits next to two other people. Let \(N\) be the number of subsets of 16 chairs that could be selected. Find the remainder when \(N\) is divided by 1000.}
} \\
\midrule
\textbf{N} & \textbf{Best-of-N (BoN)} & \textsc{\textbf{OptScale}} \\
\midrule
\endfirsthead

\toprule
\multicolumn{3}{>{\columncolor[gray]{0.95}}p{0.96\textwidth}<{\arraybackslash}}{
\textbf{(Continued) Question:} \textcolor{blue}{Sixteen chairs are arranged in a row. Eight people each select a chair in which to sit so that no person sits next to two other people. Let \(N\) be the number of subsets of 16 chairs that could be selected. Find the remainder when \(N\) is divided by 1000.}
} \\
\midrule
\textbf{N} & \textbf{Best-of-N (BoN)} & \textsc{\textbf{OptScale}} \\
\midrule
\endhead

\midrule
\multicolumn{3}{c}{\textit{Continued on next page}} \\
\bottomrule
\endfoot

\bottomrule
\endlastfoot

\textbf{Completion 3} &
\begin{minipage}[t]{\linewidth}
... To solve this problem, we model it by selecting 8 chairs such that no three selected chairs are adjacent. We can think of this as placing 8 ones among 16 positions such that no three ones are consecutive. \\
We break this into cases based on the number of blocks (runs) of 1s:
\[
\begin{aligned}
\text{For } m = 4: &\quad \binom{4}{4} \cdot \binom{9}{4} = 126 \\
m = 5: &\quad \binom{5}{3} \cdot \binom{9}{5} = 1260 \\
m = 6: &\quad \binom{6}{2} \cdot \binom{9}{6} = 1260 \\
m = 7: &\quad \binom{7}{1} \cdot \binom{9}{7} = 252 \\
m = 8: &\quad \binom{8}{0} \cdot \binom{9}{8} = 9 \\
\end{aligned}
\]
Total number of valid arrangements:
\[
126 + 1260 + 1260 + 252 + 9 = 2907
\]
Taking modulo 1000:
\[
2907 \mod 1000 = \textcolor{blue}{\boxed{907}}
\] (Correct)
\end{minipage} &
Same as BoN \\
\midrule

\textbf{Completion 6,7} &
\begin{minipage}[t]{\linewidth}
... We model the problem as binary strings of length 16 with exactly 8 ones, no three consecutive. \\
Let each run of 1s be of length 1 or 2. The remaining zeros are distributed between the runs. For each possible number of runs \(r\), compute:
\[
\binom{r}{8 - r} \cdot \binom{9}{r}
\]
Summing over \(r = 4\) to \(8\), we get:
$126 + 1260 + 1260 + 252 + 9 = 2907$ \\
$\Rightarrow 2907 \mod 1000 = \textcolor{blue}{\boxed{907}}$
\newline (Correct)
\end{minipage} &
Same as BoN \\
\midrule

\textbf{Completion 14} &
\begin{minipage}[t]{\linewidth}
(Backtracking and confusion. Model fails to converge—no final answer due to token exhaustion.)
\end{minipage} &
\textsc{OptScale} selects Completion 3 with the correct answer. \\
\midrule

\textbf{Completion 18} &
\begin{minipage}[t]{\linewidth}
... An incorrect inclusion-exclusion approach is used. Final result computed:
12870 - 15444 + 7560 = 4986  \\ $\Rightarrow 4986 \mod 1000$ = \boxed{986}

\end{minipage} &
\\
\midrule

\textbf{Completion 33, 38, 47} &
\begin{minipage}[t]{\linewidth}
... Another incorrect approach leading to:
\[
12870 - 9009 + 945 = 4806 \Rightarrow \boxed{806}
\]
\end{minipage} &
\\
\midrule

\textbf{Completion 50,51,57} &
\begin{minipage}[t]{\linewidth}
... Correctly models the constraint using inclusion-exclusion. Final count:
2907 -> \boxed{2907 \mod 1000 = 907}
\end{minipage} &
\\
\midrule

\textbf{Completion 60} &
\begin{minipage}[t]{\linewidth}
Selected Completion 18 with value \(\boxed{986}\) \\ (Incorrect answer.)
\end{minipage} &
\\
\midrule

\textbf{Token Consumption} & 545,681 & 130,593 \\
\midrule
\textbf{Correctness} & \textcolor{red}{No (\ding{55})} & \textcolor{blue}{Yes (\ding{51})} \\
\end{longtable}

\onecolumn
\begin{longtable}{p{3.5cm}|p{10cm}|p{3.5cm}}
\caption{Case 5 (Dataset: AIME25, Backbone Model: Deepseek-R1-Distill-Qwen-7B).} \label{tab:case5} \\
\toprule
\multicolumn{3}{>{\columncolor[gray]{0.95}}p{0.96\textwidth}<{\arraybackslash}}{
\textbf{Question:} \textcolor{blue}{Let \(k\) be real numbers such that the system \(|25+20i-z|=5\) and \(|z-4-k|=|z-3i-k|\) has exactly one complex solution \(z\). The sum of all possible values of \(k\) can be written as \(\frac{m}{n}\), where \(m\) and \(n\) are relatively prime positive integers. Find \(m+n\). Here \(i=\sqrt{-1}\).}
} \\
\midrule
\textbf{N} & \textbf{Best-of-N (BoN)} & \textsc{\textbf{OptScale}} \\
\midrule
\endfirsthead

\toprule
\multicolumn{3}{>{\columncolor[gray]{0.95}}p{0.96\textwidth}<{\arraybackslash}}{
\textbf{(Continued) Question:} \textcolor{blue}{Let \(k\) be real numbers such that the system \(|25+20i-z|=5\) and \(|z-4-k|=|z-3i-k|\) has exactly one complex solution \(z\). The sum of all possible values of \(k\) can be written as \(\frac{m}{n}\), where \(m\) and \(n\) are relatively prime positive integers. Find \(m+n\). Here \(i=\sqrt{-1}\).}
} \\
\midrule
\textbf{N} & \textbf{Best-of-N (BoN)} & \textsc{\textbf{OptScale}} \\
\midrule
\endhead

\midrule
\multicolumn{3}{c}{\textit{Continued on next page}} \\
\bottomrule
\endfoot

\bottomrule
\endlastfoot

\textbf{Completion 1} &
\begin{minipage}[t]{\linewidth}
\textit{Squaring both sides to remove the absolute value and square root, we obtain:}\\
\[
\left(3k + \frac{73}{2}\right)^2 = 25(2k^2 + 14k + 25)
\]
\textit{Expanding and simplifying, we get the quadratic equation:}\\
\[
164k^2 + 524k - 2829 = 0
\]
\textit{Using the quadratic formula, the sum of the roots of this quadratic equation is:}\\
\[
-\frac{b}{a} = -\frac{524}{164} = -\frac{131}{41}
\]
\textit{So, \(m=131\), \(n=41\), and \(m+n=\textcolor{red}{\boxed{172}}\).}\\
\text{(Incorrect)}
\end{minipage} &
Same as BoN \\
\midrule

\textbf{Completion 2-4} &
\begin{minipage}[t]{\linewidth}
\textit{Given \(k\) is a real number, the system:}\\
1. \(|25 + 20i - z| = 5\)\\
2. \(|z - 4 - k| = |z - 3i - k|\)\\
\textit{represents a circle and a perpendicular bisector. The bisector of points \((4+k,0)\) and \((k,3)\) has midpoint \((2+k,\tfrac{3}{2})\) and slope \(-\tfrac{3}{4}\), so the perpendicular bisector has slope \(\tfrac{4}{3}\). Equation:}\\
\[
8x - 6y - 8k - 7 = 0
\]
\textit{Distance from center \((25,20)\) to line equals radius:}\\
\[
\frac{|73 - 8k|}{10} = 5 \Rightarrow |73 - 8k| = 50
\]
\textit{Solving:}\\
\(k = \frac{23}{8}\), \(k = \frac{123}{8}\)\\
\textit{Sum: } \(\frac{146}{8} = \frac{73}{4}\), so \(m+n = \textcolor{blue}{\boxed{77}}\) \newline (Correct)
\end{minipage} &
Same as BoN \\
\midrule

\textbf{Completion 5} &
\begin{minipage}[t]{\linewidth}
\textit{Another geometric interpretation gives a different line and leads to:}\\
\[
\frac{|58k - 73|}{\sqrt{4k^2 - 32k + 100}} = 5
\]
\textit{Squaring and simplifying:}\\
\[
1088k^2 - 2556k + 943 = 0
\]
\textit{Sum of roots: } \(\frac{2556}{1088} = \frac{639}{272}\), so \(m+n = \textcolor{red}{\boxed{911}}\) \\ (Incorrect)
\end{minipage} &
Same as BoN \\
\midrule

\textbf{Completion 6} &
\begin{minipage}[t]{\linewidth}
\textit{Solving: } \(k = \frac{23}{8}\) or \(k = \frac{123}{8}\)\\
\textit{Sum: } \(\frac{146}{8} = \frac{73}{4}\), so \(m+n = \textcolor{blue}{\boxed{77}}\) \newline (Correct)
\end{minipage} &
\textsc{OptScale} terminates here with the highest verifier score, obtaining the answer 77. \\
\midrule

\textbf{Completion 22} &
\begin{minipage}[t]{\linewidth}
\textit{Incorrect values: } \(k = -\frac{23}{6}, -\frac{41}{2}\), sum \(= -\frac{73}{3}\) → \(m+n = \boxed{76}\) (Incorrect)
\end{minipage} &
\\
\midrule

\textbf{Completion 41} &
\begin{minipage}[t]{\linewidth}
\textit{Using perpendicular distance from center:}\\
\[
\frac{|8 \cdot 25 - 6 \cdot 20 - (8k + 7)|}{10} = 5 \Rightarrow |73 - 8k| = 50
\]
\textit{Solutions: } \(k = \frac{23}{8}, \frac{123}{8}\) \\
\(\Rightarrow \frac{146}{8} = \frac{73}{4}\), so \(m+n = \boxed{77}\)
\end{minipage} &
\\
\midrule

\textbf{Completion 60} &
\begin{minipage}[t]{\linewidth}
\textit{Final correct answer selected with highest verifier score: } \(\boxed{77}\)
\end{minipage} &
\\
\midrule

\textbf{Token Consumption} & 315,214 & 31,402 \\
\midrule
\textbf{Correctness} & \textcolor{blue}{Yes (\ding{51})} & \textcolor{blue}{Yes (\ding{51})}\\
\end{longtable}


\end{document}